\documentclass[acmsmall,screen]{acmart}

\AtBeginDocument{%
  }

\setcopyright{acmcopyright}
\copyrightyear{2018}
\acmYear{2018}
\acmDOI{XXXXXXX.XXXXXXX}

\usepackage{amsfonts}
\usepackage{MnSymbol}

\usepackage{bbm}

\usepackage{amsthm} 
\newtheorem{Theorem}{Theorem}  
\newtheorem{Proposition}{Proposition}
\newtheorem{Assumption}{Assumption}
\usepackage{algorithm}
\usepackage{algpseudocode}

\newcommand{\T}{\top}
\newcommand{\I}{{-1}}
\def\+#1{\mathbb{#1}}
\def\*#1{\mathbf{#1}}

\acmJournal{JACM}
\acmVolume{37}
\acmNumber{4}
\acmArticle{111}
\acmMonth{8}

\begin{document}

\title{Instrumental Variable-Driven Domain Generalization with Unobserved Confounders}

\author{Junkun Yuan}
\orcid{0000-0003-0012-7397}
\email{yuanjk@zju.edu.cn}
\affiliation{%
  \institution{Zhejiang University}
  \city{Zhejiang}
  \country{China}
}

\author{Xu Ma}
\email{maxu@zju.edu.cn}
\affiliation{%
  \institution{Zhejiang University}
  \city{Zhejiang}
  \country{China}
}

\author{Ruoxuan Xiong\textsuperscript{*}}
\email{ruoxuan.xiong@emory.edu}
\affiliation{%
  \institution{Emory University}
  \city{Atlanta}
  \country{USA}
}

\author{Mingming Gong}
\email{mingming.gong@unimelb.edu.au}
\affiliation{%
  \institution{The University of Melbourne}
  \city{Melbourne}
  \country{Australia}
}

\author{Xiangyu Liu}
\email{eason.lxy@alibaba-inc.com}
\affiliation{%
  \institution{Alibaba Group}
  \city{Zhejiang}
  \country{China}
}

\author{Fei Wu}
\email{wufei@zju.edu.cn}
\affiliation{%
  \institution{Zhejiang University,  Shanghai Institute for Advanced Study of Zhejiang University, Shanghai AI Laboratory}
  \city{Zhejiang}
  \country{China}
}

\author{Lanfen Lin}
\email{llf@zju.edu.cn}
\affiliation{%
  \institution{Zhejiang University}
  \city{Zhejiang}
  \country{China}
}

\author{Kun Kuang}
\authornote{Corresponding author.}
\email{kunkuang@zju.edu.cn}
\affiliation{%
  \institution{Zhejiang University, Key Laboratory for Corneal Diseases Research of Zhejiang Province}
  \city{Zhejiang}
  \country{China}
}


\begin{abstract}
Domain generalization (DG) aims to learn from multiple source domains a model that can generalize well on unseen target domains. Existing DG methods mainly learn the representations with invariant marginal distribution of the input features, however, the invariance of the conditional distribution of the labels given the input features is more essential for unknown domain prediction. Meanwhile, the existing of unobserved confounders which affect the input features and labels simultaneously cause spurious correlation and hinder the learning of the invariant relationship contained in the conditional distribution. Interestingly, with a causal view on the data generating process, we find that the input features of one domain are valid instrumental variables for other domains. Inspired by this finding, we propose an instrumental variable-driven DG method (IV-DG) by removing the bias of the unobserved confounders with two-stage learning. In the first stage, it learns the conditional distribution of the input features of one domain given input features of another domain. In the second stage, it estimates the relationship by predicting labels with the learned conditional distribution. Theoretical analyses and simulation experiments show that it accurately captures the invariant relationship. Extensive experiments on real-world datasets demonstrate that IV-DG method yields state-of-the-art results.
\end{abstract}



\maketitle

\section{Introduction}
General supervised learning extracts statistical patterns by assuming data across training (source) and test (target) sets are independent and identically distributed (i.i.d.). It may lead to poor generalization performance when testing the trained model on the data that is very distinct from the training one, which is known as the \emph{dataset shift} (or domain shift) problem \cite{quionero2009dataset}. A prevailing research field for addressing this problem is \emph{domain adaptation} (DA) \cite{ben2010theory}, which adapts the model from source to target with available target data. However, DA methods need to re-collect target data and repeat the model adaptation process for each new target domain, which is time-consuming or even infeasible. \emph{Domain generalization} (DG) \cite{blanchard2011generalizing} is thus proposed to use multiple semantically-related source datasets for learning a generalizable model without accessing any target data/information.

Numerous DG works \cite{li2018deep, Matsuura2020DomainGU, peng2019moment, Zhao2020DomainGV} learn domain-agnostic feature representations. Most of them \cite{li2018deep, Matsuura2020DomainGU, peng2019moment} are based on the covariate shift assumption that the marginal distribution of the input features, i.e., $P(X)$, changes yet the conditional distribution of the labels given the input features, i.e., $P(Y|X)$, stays unchanged across domains. However, it rarely holds in many real scenarios where $P(Y|X)$ also changes in different domains/environments. 
Since the goal of DG is to improve the generalization performance of the  prediction $P(Y|X)$ on unseen target domains, it is essential to capture the invariance of $P(Y|X)$ that could be extracted from the source domains.
But there might exist unobserved confounders that affect $X$ and $Y$ simultaneously. They cause spurious correlation between $X$ and $Y$, hindering the learning of the invariant relationship contained in $P(Y|X)$. 

\begin{figure*}[t]
    \centering
    \includegraphics[trim={0cm 0cm 0cm 0cm},clip,width=0.54\columnwidth]{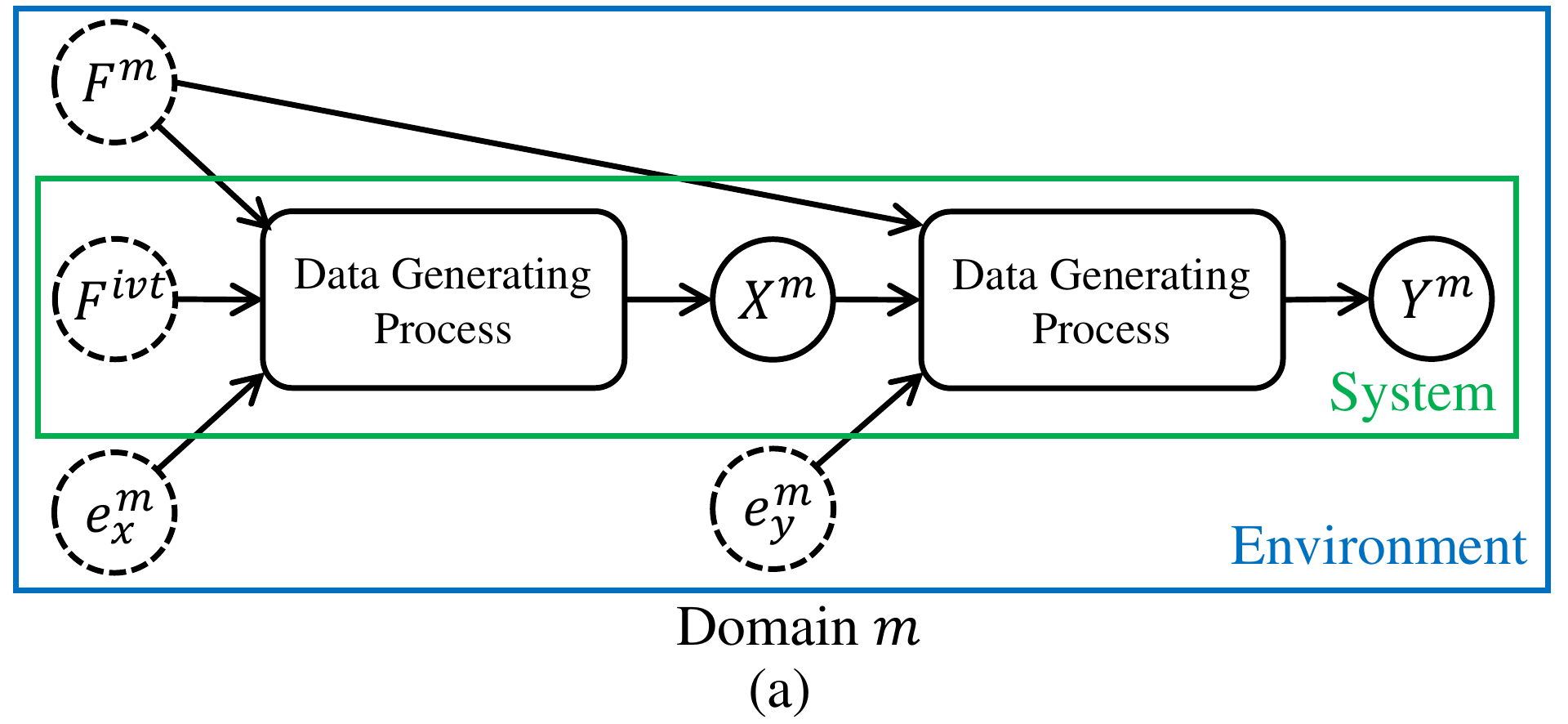}
    \includegraphics[trim={0cm 0cm 0cm 0cm},clip,width=0.44\columnwidth]{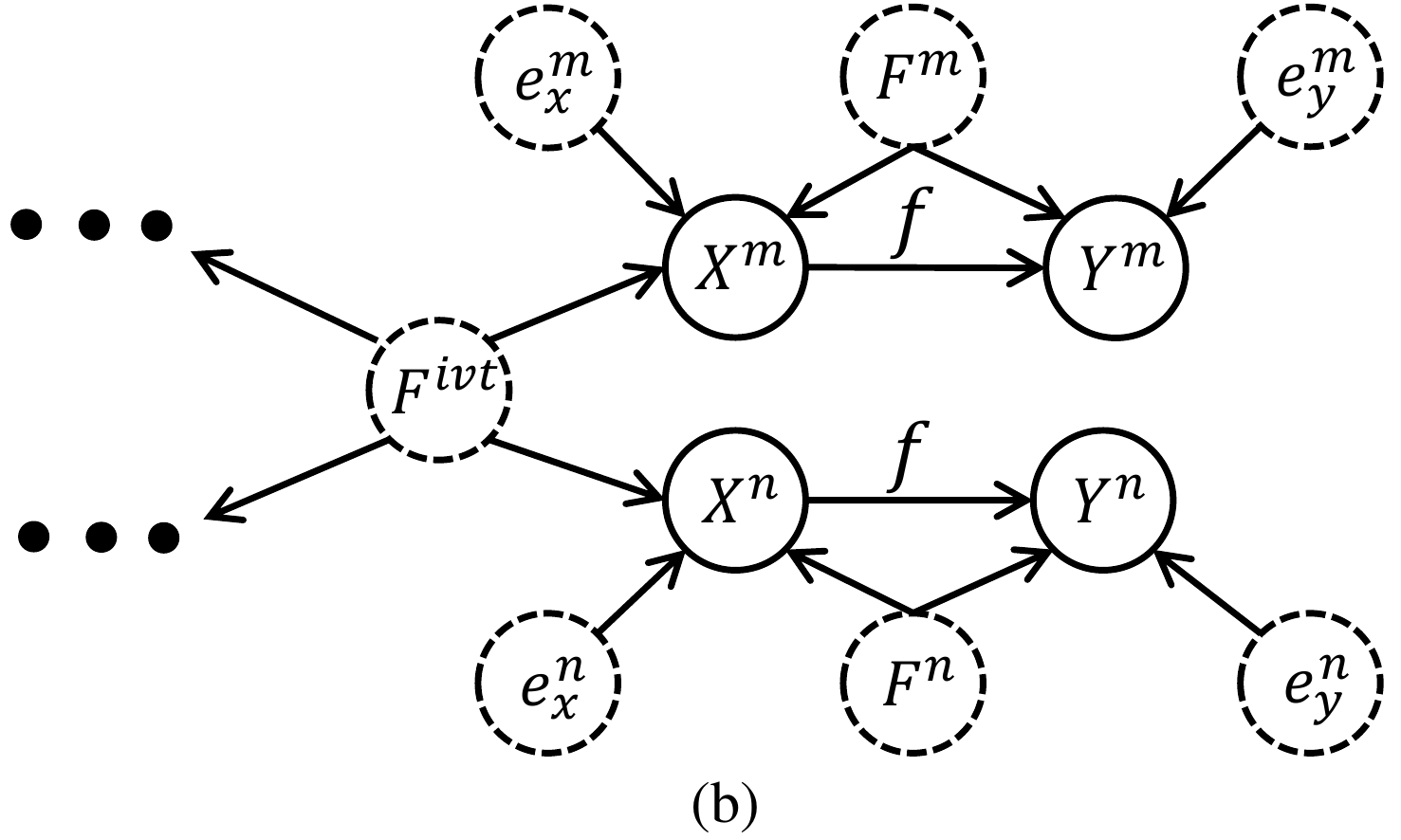}
    \caption{(a): A causal view on the data generating process for each domain $m$. (b): a causal graph for different domains. Solid and dashed circles denote observed and latent variables, respectively. For each domain $m$, input features $X^{m}$ and labels $Y^{m}$ are (indirectly) determined by domain-invariant factor $F^{ivt}$ of the system, confounded by domain-specific factor $F^{m}$ (unobserved confounder), and affected by error $e_{x}^{m}$ and $e_{y}^{m}$ from the environment. We aim to learn the invariant relationship $f$ between the input features and the labels with an instrumental variable-based method for improving the out-of-distribution generalization performance.}\label{fig-dgp-all}
\end{figure*}

In this paper, we aim to capture the invariant relationship between the input features and the labels by removing the bias of the unobserved confounders for robust domain generalization. 
We tackle this issue by putting forward a causal view on the data generating process which distinguishes domain-invariant and domain-specific parts of data as shown in Figure \ref{fig-dgp-all} (a). In an analyzed \emph{system} of domain $m$, input features $X^{m}$ and labels $Y^{m}$ are (indirectly) determined by domain-invariant factor $F^{ivt}$ that contains discriminative semantic information of objects. The domain-specific factor $F^{m}$ plays the role of a common cause of $X^{m}$ and $Y^{m}$ by affecting both of them. $X^{m}$ and $Y^{m}$ are also affected by error $e_{x}^{m}$ and $e_{y}^{m}$ from the \emph{environment} of domain $m$. In light of this, we build a causal graph of different domains in Figure \ref{fig-dgp-all} (b). We attribute the changes of the conditional distribution $P(Y|X)$ across domains to the changes of the unobserved confounder, i.e., the domain-specific factor $F^{m}$ here. Moreover, we assume that there exists an invariant relationship $f$ between the input features and the labels, contained in $P(Y|X)$, which we are interested in. 
With the analysis of Figure \ref{fig-dgp-all}, we find that the input features of one domain are valid \emph{instrumental variables} (IVs) \cite{wright1928tariff} (see Section \ref{sec-pre}) of another domain. Inspired by this finding, we propose an Instrumental Variable-driven DG method (IV-DG) to learn the relationship $f$ with two-stage learning. It first learns the conditional distribution of the input features of one domain given the input features of another domain, and then estimates $f$ by predicting the labels with the learned conditional distribution. Our method is simple yet effective that helps the model remove the bias of the unobserved confounder and learn the invariant relationship, effectively improving the generalization performance of the model. We demonstrate with theoretical analyses to verify the effectiveness of our method. Extensive experiments on both simulated and real-world data show its superior performance.

Our main contributions are summarized as follow: (i) We formulate a data generating process by distinguishing domain-invariant and domain-specific parts in data. Based on this, we build a causal graph of different domains to analyze the problem of the unobserved confounders in the domain generalization task from a causal perspective. 
(ii) We propose an Instrumental Variable-driven Domain Generalization (IV-DG) method to learn the domain-invariant relationship for improving model generalization. It exploits the input features of one domain as IVs for another domain and implements IV-based generalization learning by removing the bias of the unobserved confounders.
(iii) We provide theoretical analyses to verify our method. Moreover, extensive experiments on both simulated data and real-world datasets show that our method yields state-of-the-art results.

The remainder of this paper is organized as follows. Section \ref{sec-rel} gives a brief review of the related works on domain adaptation, domain generalization, and instrumental variables. The formulation of the investigated domain generalization problem and preliminary of instrumental variables are introduced in Section \ref{sec-pre}. The proposed method is presented in Section \ref{sec-met}. The experiments on simulated and real-world data and analysis are demonstrated in Section \ref{sec-exp}. Finally, Section \ref{sec-con} concludes this paper, with a future research outlook.

\section{Related Work}\label{sec-rel}
\subsection{Domain Adaptation and Generalization}
Domain adaptation (DA) \cite{chattopadhyay2012multisource, wu2022multiple, zhou2020progress, cai2019unsupervised, ren2020adversarial, ma2021adversarial, ma2022attention, zhang2019guide, long2015learning} aims to transfer the knowledge from the source domain(s) to the target domain(s). 
Unsupervised domain adaptation \cite{cai2019unsupervised, ma2021adversarial, zhang2019guide, long2015learning} is a prevailing direction to DA that addresses the domain shift problem by minimizing domain gap between a labeled source domain and an unlabeled target domain via domain adversarial learning \cite{cai2019unsupervised, ma2021adversarial} or domain distance minimization \cite{zhang2019guide, long2015learning}, et al. 
However, they need to access the data/information of the target domain in advance, which may be expensive or even infeasible in real scenarios. 

Domain generalization (DG) \cite{zhou2021domain, yuan2021collaborative, lv2023duet, lv2023ideal, shankar2018generalizing, volpi2018generalizing, lv2022personalizing, niu2023knowledge, yuan2023domain, yuan2022label} is proposed to use multiple labeled source domains for training a generalizable model to unseen target domains.
Recent DG methods with a variety of strategies can be included in the following main topics. The first topic is domain-invariant representation learning.
This line of works learns feature representations that are invariant to domains and discriminative for classification. Some works \cite{ghifary2015domain, li2018domain, qiao2020learning} use auto-encoder structure to obtain invariant representations by performing a data reconstruction task. Li et al. \cite{li2018domaincon} learn invariant class conditional representation with kernel mean embeddings. Piratla et al. \cite{Piratla2020EfficientDG} decompose networks into common and specific components, making the model rely on the common features. Li et al. \cite{li2018deep, Matsuura2020DomainGU} introduce adversarial learning to extract effective representations with invariance constraints. Zhao et al. \cite{Zhao2020DomainGV} introduce conditional entropy regularization term to learn conditional invariant features. Li et al. \cite{LiWW0LK20} model linear dependency in feature space and learn the common information. 
Data augmentation-based methods aim to boost generalization ability of the model by training it on various generated novel domains. 
Some methods \cite{shankar2018generalizing, volpi2018generalizing} use model gradient to perturb data and construct new datasets for model training. Some others \cite{Carlucci2019DomainGB, Wang2020LearningFE} augment datasets by solving jigsaw puzzles. Zhou et al. \cite{zhou2020deep, zhou2020learning} employ an adversarial strategy to generate novel domains while keeping semantic information consistent. A recent work \cite{zhou2021domain} mixes instance features to synthesize diverse domains and improves generalization.
Similar to the goal of DG, meta-learning-based methods keeps training the model on a meta-train dataset and improving its performance on a meta-test dataset. Numerous works \cite{balaji2018metareg, li2018learning, dou2019domain, Li2019EpisodicTF, li2019feature} put forward meta-learning guided training algorithms to improve model out-of-domain generalization. However, it may be difficult to design effective yet efficient meta-learning training algorithms in practice.
Some other methods learn the masks of features \cite{chattopadhyay2020learning} or gradient \cite{HuangWXH20} for regularization, or normalize batch/instance \cite{Seo2020LearningTO}.

\subsection{Causality-based Distribution Generalization}
To learn distribution-irrelevant features and models for stable generalization, numerous causality-based distribution generalization methods \cite{mahajan2020domain, yang2021learning, zhang2013domain, zhang2015multi, kuang2020causal, fan2020brain, miao2022domain, christiansen2021causal, gong2016domain, wu2022learning, yuan2022auto} have been introduced recently. 
For example, Yang et al. \cite{yang2021learning} investigate a robust domain adaptation problem where only a source dataset is available. They design a causal autoencoder to learn causal representations via causal structure learning.
Mahajan et al. \cite{mahajan2020domain} provide a causal interpretation of domain generalization, and show the importance of learning within-class variations for generalization. 
Lu et al. \cite{lu2022domain} and Wang et al. \cite{wang2021variational} also propose to learn domain-agnostic features for out-of-distribution generalization through knowledge distillation and variational disentanglement, respectively. 
Another direction is causal feature selection \cite{lin2022bayesian, lin2022orphicx, mao2022causal, mao2021generative}. For example, Mao et al. \cite{mao2021generative} propose to steer generative model to manufacture interventions on confounded features for learning robust visual representations.  

Compared to the previous works, our work has the following merits. 
(i) We provide a causal view on the data generating process for domain generalization with unobserved confounders. We then find that the input features can be treated as Instrumental Variables (IVs) for another domains. This finding inspires us to use the IVs to remove the domain-specific factors and capture the invariant relationship between the input features and labels. This idea of IVs for causality-based generalization learning is seldomly investigated to our knowledge, and we believe our work would shed lights on this interesting direction.
(ii) To verify this idea, we further provide theoretical insights and toy experiments which shows that the relationship estimated by our method converges to the causal invariant relationship. 
(iii) We propose a model-agnostic learning framework for the domain generalization task, which can easily deal with high-dimensional non-linear data. We implement simulation experiments on both linear and non-linear data to show the invariance learning ability of our method. Furthermore, we perform experiments on four real-world data to show the great generalization learning performance of our method. However, previous methods may either perform experiments on low-dimensional toy data \cite{zhang2013domain, gong2016domain, christiansen2021causal} or lack simulation results to show its causality learning performance \cite{mahajan2020domain, yang2021learning}.


\subsection{Instrumental Variable}
Instrumental variable (IV) method \cite{wright1928tariff} is widely employed to capture causal relationship between variables for counterfactual prediction. 
Two stage least squares (2SLS) \cite{angrist2008mostly} is the most prevailing method in IV-based counterfactual prediction, which learns $\mathbb{E}[\phi(X)|Z]$ with IV $Z$ and linear basis $\phi(\cdot)$, and fits $Y$ by least squares regression with the coefficient $\hat{\phi}(\cdot)$ estimated in the first stage.
Some non-parametric researches \cite{newey2003instrumental} extend the model basis to more complicated mapping function or regularization, e.g., polynomial basis. 
DeepIV \cite{hartford2017deep} is proposed to use deep neural networks in the two-stage procedure, it fits a mixture density network $F_{\phi}(X|Z)$ in the first stage and regresses $Y$ by sampling from the estimated mixture Gaussian distributions of $X$. 
KIV \cite{singh2019kernel} is a recent work which maps and learns the relationships among $Z$, $X$, and $Y$ in reproducing kernel Hilbert spaces. Another recent progress, DeepGMM \cite{bennett2019deep}, extends GMM methods in high-dimensional treatment and IV settings based on variational reformulation of the optimally-weighted GMM.
We follow the additive function form used by most of the previous IV-based methods, i.e., $Y=f(X)+e$.

\begin{figure*}[t]
    \centering
    \includegraphics[trim={0cm 0cm 0cm 0cm},clip,width=0.99\columnwidth]{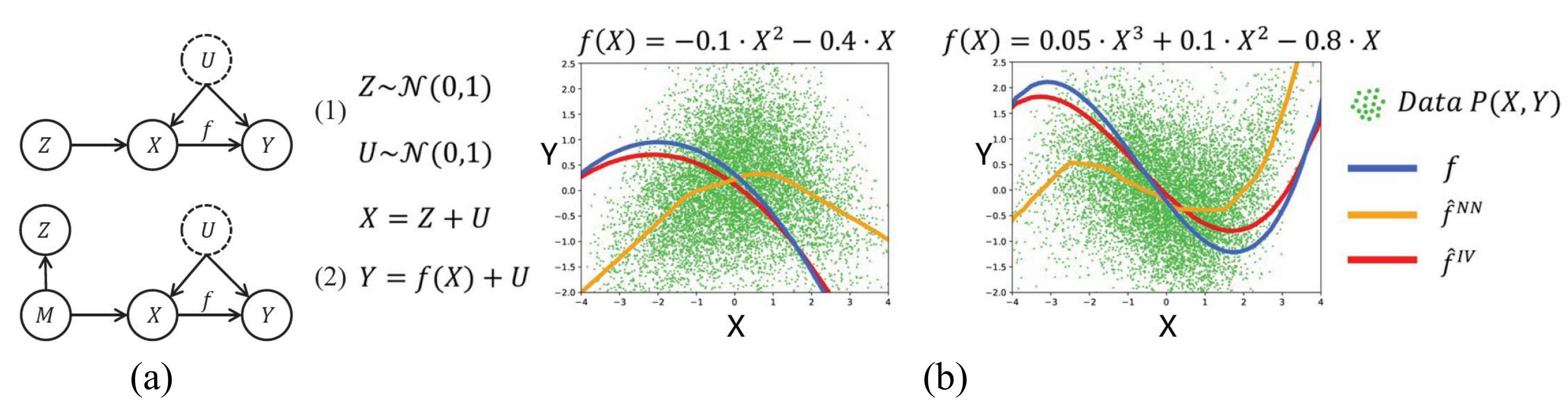}
    \caption{(a): two examples \cite{pearl2009causality} of causal structure with an instrumental variable (IV) $Z$, covariates (or input features) $X$, response (or output) $Y$, unobserved confounder $U$, and intermediate variable $M$. (b): toy experiments where $\hat{f}^{NN}$ and $\hat{f}^{IV}$ are estimated by neural networks (NN) and instrumental variable (IV) methods, respectively, on the data sampled from the biased distribution $P(X,Y)$. $\hat{f}^{NN}$ is estimated by directly taking $X$ and $Y$ as the model input and target, respectively, for training. $\hat{f}^{IV}$ is estimated via an IV-based two-stage method with IV $Z$. We utilize the IV $Z$ to estimate the invariant relationship $f$ between $X$ and $Y$ by removing the confounding effect of an unobserved confounder $U$ in the biased distribution $P(X,Y)$.}\label{fig-iv}
\end{figure*}

\section{Preliminary}\label{sec-pre}
In the domain generalization (DG) task, we have $Q$ labeled source datasets $\mathcal{D}_{1},..., \mathcal{D}_{Q}$ with different distributions $P_{1}(X^{1},Y^{1}),...,P_{Q}(X^{Q},Y^{Q})$ on joint space $\mathcal{X}\times\mathcal{Y}$, where $\mathcal{X}$ and $\mathcal{Y}$ are input feature and label spaces, respectively. In each source domain $q$, $N^{q}$ examples are sampled for the dataset $\mathcal{D}_{q}$, i.e., $\mathcal{D}_{q}=\{(\mathrm{x}_{n}^{q},y_{n}^{q})\}_{n=1}^{N^{q}}$. 
Despite the distribution shift across domains, the input features $X^q$ as well as the labels $Y^{q}$ represent the same object and used for the same task across domains.
DG aims to train a model with the $Q$ source datasets and improve its generalization performance on the unseen target domains where no data or information is provided for training.

In causal literature \cite{pearl2009causality}, invariant relationship (\emph{response function}) $f$ between covariates $X$ and response $Y$ is assumed as shown in Figure \ref{fig-iv} (a).
The unobserved confounder $U$, which causes changes to both $X$ and $Y$, introduces bias in data distribution $P(X,Y)$. The estimation of the relationship $P(Y|X)$ by learning from $P(X,Y)$ hence varies across domains with the changes of $U$. 
Instrumental variable (IV) \cite{wright1928tariff} $Z$ is a powerful tool for tracking the bias from the unobserved confounder $U$. 
A valid IV should satisfy the following conditions \cite{pearl2009causality,hartford2017deep}: (i) Relevance. $Z$ and $X$ should be relevant, i.e., $P(X|Z)\neq{P(X)}$; (ii) Exclusion. $Z$ is correlated to $Y$ only through $X$, i.e., $Z\upmodels Y|(X,U)$; and (iii) Unconfounded instrument. $Z$ is independent of $U$, i.e., $Z\upmodels U$. 
These conditions make $Z$ a valid IV, which allows us to learn the true relationship $f$ between $X$ and $Y$ by considering the changes of $Z$. 
Instead of directly leveraging $X$ to predict $Y$ for capturing relationship between $X$ and $Y$ in supervised learning, the general procedure of \textbf{two-stage} IV method is to learn the distribution of $X$ given $Z$ and then use the estimated conditional distribution to predict $Y$.

We compare the functions estimated by direct neural networks (NN), i.e., general supervised learning, and IV method by conducting toy experiments with 4000 data points sampled for training and test, respectively. 
The results are shown in Figure \ref{fig-iv} (b).
We see that NN (orange line) directly learns $P(X,Y)$ that is biased by $U$, while IV method (red line) uses $Z$ to eliminate the bias from $U$ and estimates $f$ (blue line), i.e. the invariant causal relationship, more accurately. 

For theoretical analysis, we consider a simple linear model
\begin{equation}
    \begin{aligned}
    \*Y=\*X\cdot\lambda+\*U,
    \end{aligned}
    \label{eq-pre}
\end{equation}
where $\*Y,\*U\in\mathbb{R}^{n}$, $\*X\in\mathbb{R}^{n\times d_{x}}$, $\lambda\in\mathbb{R}^{d_{x}}$, $n$ and $d_{x}$ are the number of observations and dimension of $X$, respectively. 
The invariant relationship $f$ is assumed as a linear mapping vector
$\lambda$. We estimate $\lambda$ via a two-stage IV method ($\hat{\lambda}^{IV}$) and an ordinary least squares (OLS) method ($\hat{\lambda}^{OLS}$). We have
\begin{align*}
    \hat{\lambda}^{OLS}
    =&((\*X)^{\top}\*X)^{-1}(\*X)^{\top}\*Y \\
    =&((\*X)^{\top}\*X)^{-1}(\*X)^{\top}(\*X\cdot\lambda+\*U) \\
    =&\lambda+\underbrace{((\*X)^{\top}\*X)^{-1}(\*X)^{\top}\*U}_{\text{
    not converges to 0 for $X$ is correlated with $U$
    }},
\end{align*}
\begin{align*}
    \hat{\lambda}^{IV}
    =&((\*Z)^{\top}\*X)^{-1}(\*Z)^{\top}\*Y \\
    =&((\*Z)^{\top}\*X)^{-1}(\*Z)^{\top}(\*X\cdot\lambda+\*U) \\
    =&\lambda+\underbrace{((\*Z)^{\top}\*X)^{-1}(\*Z)^{\top}\*U}_{\text{
    converges to 0 for $Z$ is independent of $U$
    }
    }.
\end{align*}
The IV method utilizes $Z$ to eliminate the bias of $U$ and the estimator $\hat{\lambda}^{IV}$ converges to $\lambda$; but the OLS estimator is biased. 
In light of this, we aim to design IV-based algorithm to capture invariant relationship between input features and labels across domains and improve the generalization performance of the trained model on unknown target domains.

\section{Instrumental Variable-Driven Generalization Learning}\label{sec-met}
We begin by giving a causal view on domain generalization. Based on it, we introduce our method, i.e., Instrumental Variable-driven Domain Generalization (IV-DG), followed by theoretical analyses. We finally demonstrate the detailed framework and algorithm of the proposed method.

\subsection{A Causal View on Domain Generalization}
The general supervised learning imposes an i.i.d. assumption, however, changes in the external environment of a new domain will lead to changes in the analyzed system (i.e., variables and their relationships). The general supervised model trained on one domain may overfit the domain-specific information, leading to the degradation of the generalization ability of the model on a new domain where the external environment changes. Nevertheless, we see that human can easily identify relationship in data no matter how the environment changes, e.g., to recognize images of animals with different backgrounds. We argue that the robust perception of human is based on the ability to distinguish domain-invariant and domain-specific parts in data via causal reasoning \cite{zhang2020domain}. In light of this, it is necessary to analyze the dataset shift problem from a causal view by defining the latent data generating process (DGP) first.

Taking a visual recognition task of animals as an example. As shown in Figure \ref{fig-dgp-all} (a), $X^{m}$ and $Y^{m}$ are images and classes sampled from a specific dataset/domain $m$. There may exist multiple causes in the DGP of $X^{m}$ and $Y^{m}$. 
Inspired by \cite{zhang2020acausal}, we argue that $X^{m}$ is determined by: (i) domain-invariant factor $F^{ivt}$, which is the key part of the recognized animals like size and limbs; (ii) domain-specific factor $F^{m}$ that changes with the external environment, like light condition and background when taking pictures; (iii) an error term $e_{x}^{m}$. 
In the DGP, the factor $F^{ivt}$ is invariant across domains, but the unobserved confounder $F^{m}$ plays the role of a common cause of $X^{m}$ and $Y^{m}$, leading to confounding bias and distribution shift across domains. For human, no matter how the images change, the corresponding classes can always be identified. It allows us to argue that there exists a latent domain-invariant relationship between input features $X^{m}$ and labels $Y^{m}$. Therefore, we let $Y^{m}$ be determined by $X^{m}$ that contains invariant information of $F^{ivt}$, and $Y^{m}$ is also affected by $F^{m}$ and $e_{y}^{m}$. Note that only the input features and labels are observed, and the others are unobserved. 

Based on the DGP, we build a causal graph of different domains as shown in Figure \ref{fig-dgp-all} (b). Input features $X^{m}$/$X^{n}$ from domain $m$/$n$ shares domain-invariant factor $F^{ivt}$ and is affected by different domain-specific factor $F^{m}$/$F^{n}$ and error $e_{x}^{m}$/$e_{x}^{n}$. Label $Y^{m}$/$Y^{n}$ is determined by $X^{m}$/$X^{n}$ through the relationship $f$, and is influenced by $F^{m}$/$F^{n}$ and $e_{y}^{m}$/$e_{y}^{n}$.
\begin{Assumption}\label{assumption-data}
    Data distributions of different domains satisfy the data generating process and causal graph in Figure \ref{fig-dgp-all}, where only the factor $F^{ivt}$ and relationship $f$ are invariant.
\end{Assumption}
In each domain $m$, general supervised learning trains the model to learn conditional distribution:
\begin{equation}
    \begin{aligned}
        P(Y^{m}|X^{m})=&\int{P(Y^{m}|X^{m},F^{m})P(F^{m}|X^{m})}dP(F^{m}).
    \end{aligned}
\end{equation}
The domain-specific factor $F^{m}$ is a common cause of $X^{m}$ and $Y^{m}$, leading to spurious correlation between $X^{m}$ and $Y^{m}$, hence the conditional distribution changes across domains.
Since $F^{m}$ is latent and can not be controlled, the introduced bias in data may not be removed directly. The model trained by minimizing risk on one domain overfits the bias and may have terrible performance on a new domain where the spurious correlation is different with the changes of the domain-specific factor.
Since directly minimizing the risk on target domains is impossible as the data is unknown, instead, we propose to learn the relationship $f$ between the input features and the labels which is invariant across domains. In causal literature \cite{pearl2009causality}, utilizing instrumental variable (IV) is an effective way to address the spurious correlation from the unobserved factor. By finding that the input features of one domain are valid IVs for other domains, we propose an IV-based two-stage method to learn the relationship $f$ for stable domain generalization, which is introduced in the following. 

\subsection{Learning Domain-Invariant Relationship with Instrumental Variable}
Under Assumption \ref{assumption-data}, we give the following conclusions by using d-separation criterion \cite{pearl2009causality}.
\begin{Proposition}\label{d-separation}
    For any two domains $m$ and $n$, if $m\neq{n}$, then the following conditions hold: (1) $X^{n}\nupmodels{X^{m}}$; (2) $X^{n}\upmodels{Y^{m}}|(X^{m},F^{m})$; (3) $X^{n}\upmodels{F^{m}}$; and (4) $X^{n}\upmodels{e_{y}^{m}}$.
\end{Proposition}
Based on the above proposition, we have the following finding. 
\begin{Theorem}\label{the-iv}
    For any two domains $m$ and $n$, if $m\neq{n}$, then $X^{n}$ is a valid instrumental variable of domain $m$.
\end{Theorem}
This theorem can be proved by referring to the conditions of IV in Section \ref{sec-pre}. It indicates that one may adopt the input features of one source dataset as valid IVs to estimate the domain-invariant relationship $f$ with another source dataset via a two-stage IV process (see Section \ref{sec-pre}). 
That is, we first estimate the conditional distribution $P(X^{m}|X^{n})$, and then predict labels $Y^{m}$ with $P(X^{m}|X^{n})$ instead of the input features $X^{m}$. Since $X^{n}$ is independent of $F^{m}$, the changes of $X^{n}$ through $X^{m}$ to $Y^{m}$ is stable to the changes of $F^{m}$. The estimation process can be understood as indirectly learning the changes of $X^{m}$ with the changes of $F^{ivt}$, i.e., parent of $X^{m}$ and $X^{n}$. $F^{ivt}$ determines the class of the analyzed system, hence the estimated relationship between $X^{m}$ and $Y^{m}$ via this two-stage procedure is discriminative for classification yet insensitive to domain changes. 
By following \cite{hartford2017deep, singh2019kernel, bennett2019deep}, we assume that the label $Y^{m}\in\mathbb{R}$ is structurally determined by the following DGP:
\begin{equation}\label{equ-dgp}
    Y^{m}=f(X^{m})+\alpha_m^\T F^{m}+e_{y}^{m},
\end{equation}
where $f(\cdot)$ is an unknown continuous function, $\alpha_m\in\mathbb{R}^{d_{f}}$ is coefficient vector of $F^m\in\mathbb{R}^{d_{f}}$, $d_f$ is the dimension of factor, $\mathbb{E}[F^{m}]=\*0$ and $\mathbb{E}[e_{y}^{m}]=0$.
By taking the expectation of $Y^{m}$ conditional on $X^{n}$, we have:
\begin{equation}
    \begin{aligned}
    \mathbb{E}[Y^{m}|X^{n}]=&\mathbb{E}[f(X^{m})|X^{n}]+\underbrace{\mathbb{E}[\alpha_m^\T F^{m}|X^{n}]+\mathbb{E}[e_{y}^{m}|X^{n}]}_{=0 \text{ for } X^{n}\upmodels{F^{m}} \text{ and } X^{n}\upmodels{e_{y}^{m}}} \\
    =&\underbrace{\int{f(X^{m})\underbrace{dP(X^{m}|X^{n})}_{\text{stage 1}}}}_{\text{stage 2}}.
    \end{aligned}
\end{equation}
It yields a two-stage strategy of learning the invariant relationship $f$ with the instrumental variable $X^{n}$. That is, in the first stage, we estimate conditional distribution $P(X^{m}|X^{n})$; and in the second stage, we estimate the invariant relationship $f$ via the approximation of $P(X^{m}|X^{n})$ learned in the first stage, i.e., predict the label $Y^m$ with the estimated $P(X^{m}|X^{n})$.

We further consider a linear setting to make it clearer. Let the dimensions of the factors and input features be $d_{f}$ and $d_{x}$, respectively, i.e., $F^{ivt}\in\mathbb{R}^{d_{f}}$, $F^{m}\in\mathbb{R}^{d_{f}}$, $X^{m}\in\mathbb{R}^{d_{x}}$. Note that we assume $F^{ivt}$ and $F^{m}$ have the same dimension, because they could be the extracted features from data, as implemented in our framework. The error terms and label are real numbers, i.e., $e_{x}^{m}\in\mathbb{R}$, $e_{y}^{m}\in\mathbb{R}$, $Y^{m}\in\mathbb{R}$. 
Assume that we sample $n$ observations from each domain. We stack all observations together, i.e., let $\*X^{m}$ be the matrix where $i$-th row is observation $(x_{i}^{m})^\T$. Other bold symbols are similarly defined. The DGP is then assumed as:
\begin{equation}
    \begin{aligned}
    \mathbf{X}^m=& \begin{bmatrix} \mathbf{F}^{ivt} & \mathbf{F}^{m} \end{bmatrix}  \begin{bmatrix} \boldsymbol{\phi}_{m} \\ \boldsymbol{\alpha}_{m}  \end{bmatrix}   + \mathbf{e}^m_{x}\\
    \mathbf{Y}^m =& \begin{bmatrix} \mathbf{X}^m & \mathbf{F}^{m} \end{bmatrix} \begin{bmatrix} \lambda_{ivt} \\ \beta_m \end{bmatrix} + \mathbf{e}^m_{y},
\end{aligned}\label{eq-xy-m}
\end{equation}
where $\boldsymbol{\phi}_{m}$, $\boldsymbol{\alpha}_{m}$, $\lambda_{ivt}$, and $\beta_{m}$ are coefficients. Note that $\lambda_{ivt}$ is the invariant relationship between input features and labels. Let input features $X^n$ from domain $n$, where $n\neq{m}$, be the IV for performing the two-stage IV method.
The first stage is to learn the conditional distribution of $X^m$ by regressing 
$\mathbf{X}^m$ on the IV $\mathbf{X}^n$ with $\hat\gamma$, that is,
\begin{equation}
    \begin{aligned}
        \hat\gamma =& \big( (\mathbf{X}^n)^{\top} \mathbf{X}^n \big)^{-1} (\mathbf{X}^n)^{\top} \mathbf{X}^m.
    \end{aligned}
\end{equation}
Then, the second stage is to predict label $\*Y^m$ with the estimated conditional distribution, i.e., regressing $\*Y^m$ on $\hat{\*X}^m=\*X^n\hat\gamma$ with estimated relationship $\hat{\lambda}_{ivt}^{IV}$, that is,
\begin{equation}
    \begin{aligned}
            \hat{\lambda}_{ivt}^{IV} =& \big( (\hat{\mathbf{X}}^m)^{\top} \hat{\mathbf{X}}^m \big)^{-1} (\hat{\mathbf{X}}^m)^{\top} \mathbf{Y}^m.
    \end{aligned}
\end{equation}
Here, $\mathbf{Y}^{m}$, $\mathbf{X}^{m}$, and $\lambda_{ivt}$ in Eq. (\ref{eq-xy-m}) correspond to $\*Y$, $\*X$, and $\lambda$ in Eq. (\ref{eq-pre}), respectively. $\mathbf{F}^{m}\beta_m$ in Eq. (\ref{eq-xy-m}) corresponds to the unobserved confounder $\*U$ in Eq. (\ref{eq-pre}). Different from the preliminary section that the IV $\*Z$ is available, we consider a more practical scenario that only the input features and labels are available. Thus, we propose to utilize the input features of another domain, i.e., $\*X^{n}$, as IV to perform the two-stage learning process introduced in the preliminary section.

Then, we have the following theorem.
\begin{Theorem}\label{the-iv}
    Suppose the minimum eigenvalue of $\boldsymbol{\phi}_{m}^\T \cdot \+E[F^{ivt}(F^{ivt})^\T] \cdot \boldsymbol{\phi}_{m}$ is bounded away from 0, and each variable of $F^{ivt}$, $F^{m}$, $e_x^n$, and $e_y^n$ of a random domain $m$ has a finite variance, then $\hat{\lambda}_{ivt}^{IV}$ is a consistent estimator which converges to $\lambda_{ivt}$, that is, $\hat{\lambda}_{ivt}^{IV} = {\lambda}_{ivt} + O_{p}\left(\frac{1}{\sqrt{n}}\right)$.
\end{Theorem}

\emph{Proof.} Since $F^{m}$ is uncorrelated with $F^{ivt}$, $F^{n}$, and $e_{x}^{n}$ (d-separation), together with $\+E[F^m] = \*0$ and $\+E[e^n_x] = 0$, we have
\begin{equation}
    \begin{aligned}
        \frac{1}{n}   (\*F^{ivt})^\T \mathbf{F}^m = O_p\bigg( \frac{1}{\sqrt{n}} \bigg), \\
        \frac{1}{n}   (\*F^n)^\T \mathbf{F}^m = O_p\bigg( \frac{1}{\sqrt{n}} \bigg), \\
  \frac{1}{n}   (\*e^n_x)^\T \mathbf{F}^m = O_p\bigg( \frac{1}{\sqrt{n}} \bigg).
    \end{aligned}\label{Eq: clt}  
\end{equation}

Eq. (\ref{Eq: clt}) can be proved based on Central Limit Theorem (CLT) \cite{heyde2006central}. Specifically, we assume two uncorrelated variables $x_a$ and $x_b$ with zero means and finite variances, and let $x=x_a x_b$. Using CLT, we know that as the sample numbers $n$ becomes large, the distribution of $\sqrt{n}\bar{x}$ converges in distribution to a normal distribution with mean $0$ and variance $\sigma^2$, where $\sigma^2$ is the variance of $x$. That is, $\sqrt{n}\bar{x} \xrightarrow{d} \mathcal{N}(0,\sigma^2)$, which can be rewritten to $\frac{\bar{x}}{\sqrt{\frac{\sigma^2}{n}}} \xrightarrow{d} \mathcal{N}(0,1)$. Since $\sigma^2$ is the variance of $x$, we have $\sqrt{\frac{\sigma^2}{n}} = \frac{\sigma}{\sqrt{n}}$. Then, $\frac{\bar{x}}{\sigma/\sqrt{n}} \xrightarrow{d} \mathcal{N}(0,1)$. This implies that $\frac{\bar{x}}{\sigma/\sqrt{n}} = O_p(1)$, which in turn implies that $\bar{x} = O_p(\frac{\sigma}{\sqrt{n}})$. Since $\sigma$ is a constant (dependent on the distribution of $x_i$), we have $\bar{x} = O_p(\frac{1}{\sqrt{n}})$. Hence, $1/n \sum_i x_i = \bar{x} = O_p(1/\sqrt{n})$. Therefore, Eq. (\ref{Eq: clt}) holds.

Then, we have
\begin{equation}\label{eq:xn-fm}
    \begin{aligned}
        \frac{1}{n} (\mathbf{X}^n)^{\top} \mathbf{F}^m 
        =& \frac{1}{n} \big( \*F^{ivt} \boldsymbol{\phi}_{n} + \*F^{n} \boldsymbol{\alpha}_n + \mathbf{e}^n_{x} \big)^\T \mathbf{F}^m
        = O_p\bigg( \frac{1}{\sqrt{n}} \bigg).
    \end{aligned}
\end{equation}
Similarly, since $e_{y}^{m}$ is independent of $F^{ivt}$, $F^{n}$, and $e_{x}^{n}$, then
\begin{equation}
    \begin{aligned}
        \frac{1}{n} (\mathbf{X}^n)^{\top} \mathbf{e}_y^m 
        =& \frac{1}{n_m} \big( \*F^{ivt} \boldsymbol{\phi}_{n} + \*F^{n} \boldsymbol{\alpha}_n + \mathbf{e}^n_{x} \big)^\T \mathbf{e}_y^m
        = O_p\bigg( \frac{1}{\sqrt{n}} \bigg).
    \end{aligned}
\end{equation}
We then have
\begin{equation}
    \begin{aligned}
        \frac{1}{n} (\mathbf{X}^m)^{\top} \mathbf{X}^n  
        =& \frac{1}{n}  \big( \*F^{ivt} \boldsymbol{\phi}_{m} + \*F^{m} \boldsymbol{\alpha}_{m} + \mathbf{e}^m_{x}   \big)^\T  \\
        &\cdot \big( \*F^{ivt} \boldsymbol{\phi}_{n} + \*F^{n} \boldsymbol{\alpha}_{n} + \mathbf{e}^n_{x}  \big) \\
        =& \frac{1}{n} \boldsymbol{\phi}_{m}^\T (\*F^{ivt})^\T \*F^{ivt} \boldsymbol{\phi}_{n} + O_p\bigg( \frac{1}{\sqrt{n}} \bigg),
    \end{aligned}
\end{equation}
\begin{equation}\label{equation-xt-xt}
    \begin{aligned}
    \frac{1}{n} (\mathbf{X}^n)^{\top} \mathbf{X}^n 
    =& \frac{1}{n}  \big( \*F^{ivt} \boldsymbol{\phi}_{n} + \*F^{n} \boldsymbol{\alpha}_n + \mathbf{e}^n_{x}   \big)^\T \cdot \big( \*F^{ivt} \boldsymbol{\phi}_{n} + \*F^{n} \boldsymbol{\alpha}_{n} + \mathbf{e}^n_{x}  \big) \\
    =& \frac{1}{n} \left(\boldsymbol{\phi}_{n}^\T (\*F^{ivt})^\T \*F^{ivt} \boldsymbol{\phi}_{n}+\boldsymbol{\alpha}_{n}^{\top}(\*F^{n})^\T \*F^{n}\boldsymbol{\alpha}_{n}+(\*e_{x}^{n})^{\top}\*e_{x}^{n}\right.+\left. O_p\left( \frac{1}{\sqrt{n}} \right)\right).
\end{aligned}
\end{equation}
Note that $\boldsymbol{\alpha}_{n}^{\top}(\*F^{n})^{\top}\*F^{n}\boldsymbol{\alpha}_{n}/n$ and $(\*e_{x}^{n})^{\top}\*e_{x}^{n}/n$ are positive semi-definite matrices and the minimum eigenvalue of $\boldsymbol{\phi}_{n}^{\T}\cdot\+E[F^{ivt}(F^{ivt})^{\T}]\cdot\boldsymbol{\phi}_{n}$ is bounded away from 0. Hence, the minimum eigenvalue of $\boldsymbol{\phi}_{n}^\T \cdot \+E[F^{ivt}(F^{ivt})^\T] \cdot \boldsymbol{\phi}_{n} + \boldsymbol{\alpha}_{n}^\T \cdot \+E[F^{n}(F^{n})^\T] \cdot \boldsymbol{\alpha}_{n} + \+E[e_{x}^{n} (e_{x}^{n})^\T]$ is bounded away from 0, then
\begin{equation}\label{equation-reverse-fivt-ft-ext}
    \begin{aligned}
        &\bigg(\frac{1}{n} \bigg(\boldsymbol{\phi}_{n}^\T (\*F^{ivt})^\T \*F^{ivt} \boldsymbol{\phi}_{n}+\boldsymbol{\alpha}_{n}^{\top}(\*F^{n})^{\top}\*F^{n}\boldsymbol{\alpha}_{n}+(\*e_{x}^{n})^{\top}\*e_{x}^{n}+ O_p\bigg( \frac{1}{\sqrt{n}} \bigg)\bigg)\bigg)^{-1}\\
        =&\bigg(\boldsymbol{\phi}_{n}^\T \cdot \+E[F^{ivt}(F^{ivt})^\T] \cdot \boldsymbol{\phi}_{n}+ \boldsymbol{\alpha}_{n}^\T \cdot \+E[F^{n}(F^{n})^\T] \cdot \boldsymbol{\alpha}_{n} + \+E[e_{x}^{n} (e_{x}^{n})^\T ]  \bigg)^{-1}+O_p\bigg( \frac{1}{\sqrt{n}} \bigg).
    \end{aligned}
\end{equation}
Therefore, by Eq. (\ref{eq:xn-fm}-\ref{equation-reverse-fivt-ft-ext}),
\begin{align*}
    \hat{\lambda}_{ivt}^{IV} =& \big( (\hat{\mathbf{X}}^m)^{\top} \hat{\mathbf{X}}^m \big)^{-1} (\hat{\mathbf{X}}^m)^{\top} \mathbf{Y}^m \\
    =& \bigg( (\mathbf{X}^m)^{\top} \mathbf{X}^n \big( (\mathbf{X}^n)^{\top} \mathbf{X}^n \big)^{-1}   (\mathbf{X}^n)^{\top} \mathbf{X}^m \bigg)^\I  \\ 
    & \cdot (\mathbf{X}^m)^{\top} \mathbf{X}^n \big( (\mathbf{X}^n)^{\top} \mathbf{X}^n \big)^{-1} (\mathbf{X}^n)^{\top}\\
    &\cdot \big(\mathbf{X}^m \lambda_{ivt}  +  \mathbf{F}^m\beta_{m} + \mathbf{e}^m_{y} \big)  \\
    =&\lambda_{ivt} +  O_p\bigg( \frac{1}{\sqrt{n}} \bigg).
\end{align*}
This theorem indicates that the coefficient $\hat{\lambda}^{IV}_{ivt}$ estimated by the two-stage IV method is a consistent estimator that converges to $\lambda_{ivt}$, which is the invariant relationship between input features and label. In light of this, we propose our method IV-DG which can capture invariant relationship and yield stable generalization performance even on high-dimensional real-world data.

\begin{figure*}[t]
    \centering
    \includegraphics[trim={0cm 0cm 0cm 0cm},clip,width=0.9\columnwidth]{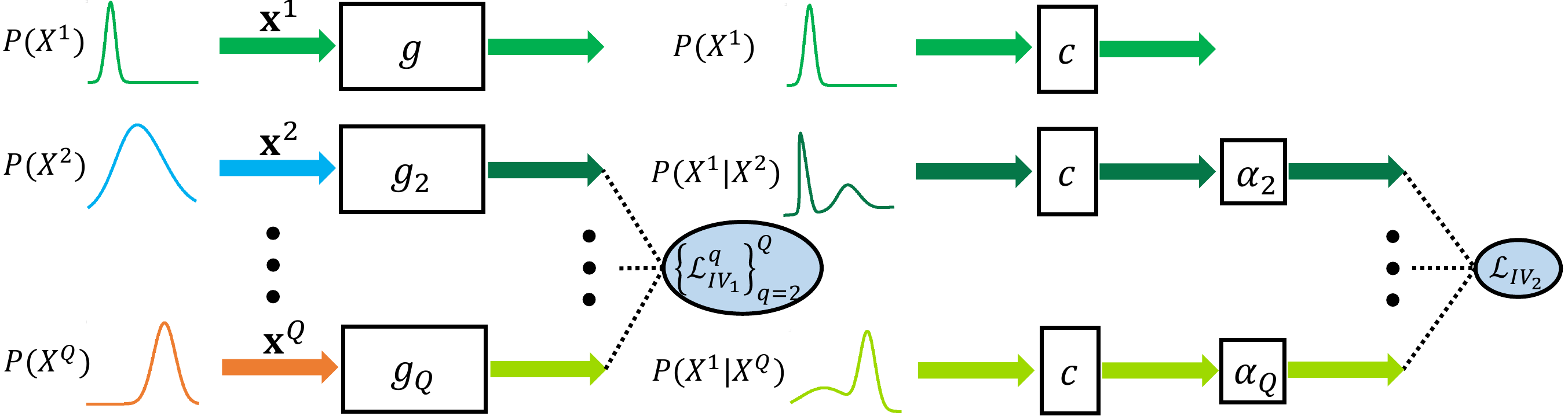} 
    \caption{The proposed IV-DG framework. Feature extractor $g$ extracts input features of data and the networks $g_{2},...,g_{Q}$ learns conditional distributions $P(X^1|X^2),...,P(X^1|X^Q)$, respectively. We learn invariant relationship between input features and label with a linear classifier $c$ via an IV-based two-stage method. It learns conditional distributions of $X^1$ given IV $X^2,...,X^Q$ with loss $\{\mathcal{L}_{IV_{1}^{q}}\}_{q=2}^{Q}$ by optimizing $g_{2},...,g_{Q}$, then uses the estimated conditional distributions to optimize $c$ by predicting $Y^{1}$ with loss $\mathcal{L}_{IV_{2}}$. Hyper-parameters $\{\alpha_{q}\}_{q=2}^{Q}$ tune the influence of each IV, i.e., $X^2,...,X^Q$, during training.}
    \label{fig-framework}
\end{figure*}

\subsection{Framework and Algorithm}
\begin{algorithm}[t]
    \caption{Instrumental Variable-driven Domain Generalization (IV-DG)}  
    \label{alg}
    \begin{algorithmic}[1]
        \Require  
        Datasets $\mathcal{D}_{1},...,\mathcal{D}_{Q}$, mixed dataset $\mathcal{D}_{mix}$, batchsize $B$, epochs $E^{pre}$, $E^{IV}$;
    \Ensure  
        Well-trained $\hat{g}$ and $\hat{c}$;
    \For{$epoch=1$ to $E^{pre}$} {$\quad$ // model pretraining}
        \State Sample $B$ examples from $\mathcal{D}_{mix}$ and optimize $g$, $c$ by minimizing $\mathcal{L}_{pre}$ as Eq. (\ref{equ-pre});
    \EndFor
    \State Initialize $g_{q}$ by $g_{q}\leftarrow{g}$ for each $q\in\{2,...,Q\}$;
    \For{$epoch=1$ to $E^{IV}$} $\quad$ // a two-stage IV method 
        \For{$q=2$ to $Q$}
            \State Sample $B$ examples from $\mathcal{D}_{1}$ and $\mathcal{D}_{q}$ and optimize $g_{q}$ by minimizing $\mathcal{L}_{IV_{1}^{q}}$ as Eq. (\ref{equ-iv1});
        \EndFor
        \State Sample $B$ examples from $\mathcal{D}_{q}$, $q=2,...,Q$, and optimize $c$ by minimizing $\mathcal{L}_{IV_{2}}$ as Eq. (\ref{equ-iv2}).
    \EndFor
  \end{algorithmic}
\end{algorithm}

Based on our analysis, we propose our method IV-DG with framework and algorithm as shown in Figure \ref{fig-framework} and Algorithm \ref{alg}, respectively. By following the common framework of DG, we adopt a feature extractor (backbone) $g$ which extracts the input features of high-dimensional data. We exploit a linear classifier $c$ to capture the invariant relationship between input features and label. We first pretrain $g$ and $c$ with mixed source data to initialize the ability of feature extraction and prediction, respectively, and then perform an IV-based two-stage method to debias $c$ for boosting the generalization ability of the model. Note that we can randomly select one source domain to be domain 1, while the other source domains to be domain 2,...,Q. We learn the invariant relationship between input features and label, i.e., an debiased classifier $c$, on domain 1, when the input features of the other domains are used as IVs.

We first pretrain the feature extractor $g$ and the classifier $c$ to initialize their ability of feature extraction and prediction, respectively. We randomly mix the sources $\{\mathcal{D}_{q}\}_{q=1}^{Q}$ to build a mixed dataset $\mathcal{D}_{mix}$ and use it to pretrain $g$ and $c$ with a cross-entropy classification loss $\mathcal{L}_{pre}$:
\begin{equation}\label{equ-pre}
    \mathcal{L}_{pre}={\mathbb{E}_{(\mathrm{x},y)\in\mathcal{D}_{mix}}}\ell\left(c\circ{g}\left(\mathrm{x}\right),y\right),
\end{equation}
where $\ell$ is the cross-entropy loss function. Through model pretraining, the feature extractor $g$ learns to extract feature representations of different datasets, and the classifier $c$ is initialized to classify the extracted feature representations. However, $c$ is biased because of the domain-specific information from the source datasets. We then perform an IV-based two-stage method to debias $c$ for learning the invariant relationship between the input feature (representations) and the labels.

\textbf{Remark.} Note that bias from source domains could be brought in this process, which may affect the learning of IV method in the following process. Despite this, this representation learning process could effectively help us extract features of each source domain for performing IV method. We assume that the introduced error will not have a significant impact on the final results.

In the stage 1 of the IV method, we assign the parameters of $g$ to $\{g_{q}\}_{q=2}^{Q}$ to initialize them, which is effective for the first stage of the IV method to our empirical experience. The two-stage IV method is conducted by: (i) learning conditional distributions of $X^{1}$ given IV $X^{2},...,X^{Q}$ via optimizing the networks $\{g_{q}\}_{q=2}^{Q}$; (ii) using the learned conditional distributions, i.e., $P(X^1|X^2),...,P(X^1|X^Q)$, to optimize the classifier $c$ by predicting $Y^1$.
Specifically, for the first stage, we use $g_{q}$ to estimate $P(X^{1}|X^{q})$ with the Maximum Mean Discrepancy (MMD) \cite{gretton2012kernel}, i.e., $d_{k}^{2}(v,w)\triangleq\Vert\mathbb{E}_{v}[\phi(g_{q}(\mathrm{x}^{q}))]-\mathbb{E}_{w}[\phi(g(\mathrm{x}^{1}))]\Vert_{\mathcal{H}_{k}}^{2}$.
The distributions of the extracted input feature representations $g_{q}(\mathrm{x}^{q})$ and $g(x^{1})$, i.e., $v$ and $w$, satisfy $v=w$ iff $d_{k}^{2}(v,w)=0$. A characteristic kernel $k(g_{q}(\mathrm{x}^{q}),g(\mathrm{x}^{1}))=<\phi(g_{q}(\mathrm{x}^{q})),\phi(g(\mathrm{x}^{1}))>$ is defined as a convex combination of $o$ positive semi-definite kernels $\{k_{u}\}$, i.e., $\mathcal{K}\triangleq\left\{k=\sum_{u=1}^{o}\beta_{u}k_{u}:\sum_{u=1}^{o}\beta_{u}=1,\beta_{u}>=0,\forall{u}\right\}$, where $\beta_{u}$ guarantees the characteristic of multi-kernel $k$ \cite{long2015learning, long2017deep}. We then estimate $P(X^{1}|X^{q})$ by optimizing $g_{q}$, which minimizes the MMD distance between the feature representations of $X^{1}$ and $X^{q}$ with the loss function
\begin{equation}\label{equ-iv1}
    \begin{aligned}
        \mathcal{L}_{IV_{1}}^{q}
        =&p_{q,1}d_{k}^{2}\left(g_{q}\left(\mathrm{x}^{q}\right),g\left(\mathrm{x}^{1}\right)\right).
    \end{aligned}
\end{equation}
where $p_{q,1}:=\mathbb{I}(y^{q}=y^{1})$, i.e., $p_{q,1}=1$ when $y^{q}=y^{1}$, otherwise $p_{q,1}=0$. It is used to guarantee only the MMD distance of the input features from the same classes are minimized, which helps $g_{q}$ to learn a more accurate conditional distribution $P(X^{1}|X^{q})$ for each $q\in\{2,...,Q\}$. 

In the stage 2 of the IV method, we sample points from the conditional distributions estimated in the first stage and use them to predict the labels. We optimize the classifier $c$ with a classification loss of the estimated conditional distribution, that is,
\begin{equation}\label{equ-iv2}
    \mathcal{L}_{IV_{2}}=\frac{1}{Q-1}\sum_{q=2}^{Q}\alpha_{q}\mathbb{E}_{(\mathrm{x}^{q},y^{q}),(\mathrm{x}^{1},y^{1})}\left[p_{q,1}\ell\left(c\circ g_{q}\left(\mathrm{x}^{q}\right),y^{1}\right)\right]
\end{equation}

Since input features $X^{q}$ of each domain $q$, where $q\in\{2,...,Q\}$, could be used as an IV to capture the invariant relationship, we set hyper-parameters $\{\alpha_{q}\}_{q=2}^{Q}$ to tune the influence of each IV in the learning process for further improving model generalization. 
We use $p_{q,1}$ to guarantee the data used for debiasing are in the same classes. By optimizing $\mathcal{L}_{IV_{2}}$, the classifier $c$ removes the domain-specific bias of the source datasets introduced in the model pretraining. It allows $c$ to capture the domain-invariant relationship, improving the out-of-domain generalization ability. 

\textbf{Remark.} Since we adopt $p_{q,1}$ to align the labels of domain 1 and domain $q$ ($q\in\{2,...,Q\}$), i.e., $Y^1$ and $Y^q$, at the two stages of IV-DG, IV process is implemented for each class during training. For example, when we sample data points from one class in domain $m$, we then sample data points from the same class in domain $n$ for learning (because we make $Y^m=Y^n$ in the training process). Even if there is not a triple of treatment $X^m$, instrument $X^n$, and outcome $Y^m$, we make $X^m$ and $X^n$ connected by letting $Y^m=Y^n$. Because $X^m$ and $X^n$ share the factor $F^{ivt}$ which is domain-invariant for a specific class when $Y^m=Y^n$. From another perspective, please see the Cats and Dogs dataset introduced in Section \ref{sec:biased-data}. Domain TB1 contains bright dogs and dark cats but domain TB2 contains dark dogs and bright cats. If we train a model on the domain TB1 or the domain TB2, the model would be biased by the brightness of the animals. In our IV-based method IV-DG, we can make $Y^m=Y^n=cat$ and let $X^m$ and $X^n$ be the cat features sampled from the domain TB1 and TB2, respectively, then capture the invariant relationship between the cat features and its label. Because the cat features from different domains share the factor $F^{ivt}$, which is the invariant characteristics of cats, like the shape of cats. The same process is performed for the dogs, too. Although we may not get triple samples from real-world datasets, we argue that our method still can capture stable relationship between input features and labels under the strong learning ability of deep neural networks, as shown by extensive experiments.

\section{Experiments}\label{sec-exp}
We first conduct simulation experiments to verify the relationship learned by our method IV-DG. Then, we perform experiments on multiple real-world datasets to further testify the model generalization performance achieved by IV-DG.

\subsection{Experiments on Simulated Datasets}

\begin{figure*}[t]
    \centering
    \includegraphics[trim={0cm 0cm 0cm 0cm},clip,width=0.99\columnwidth]{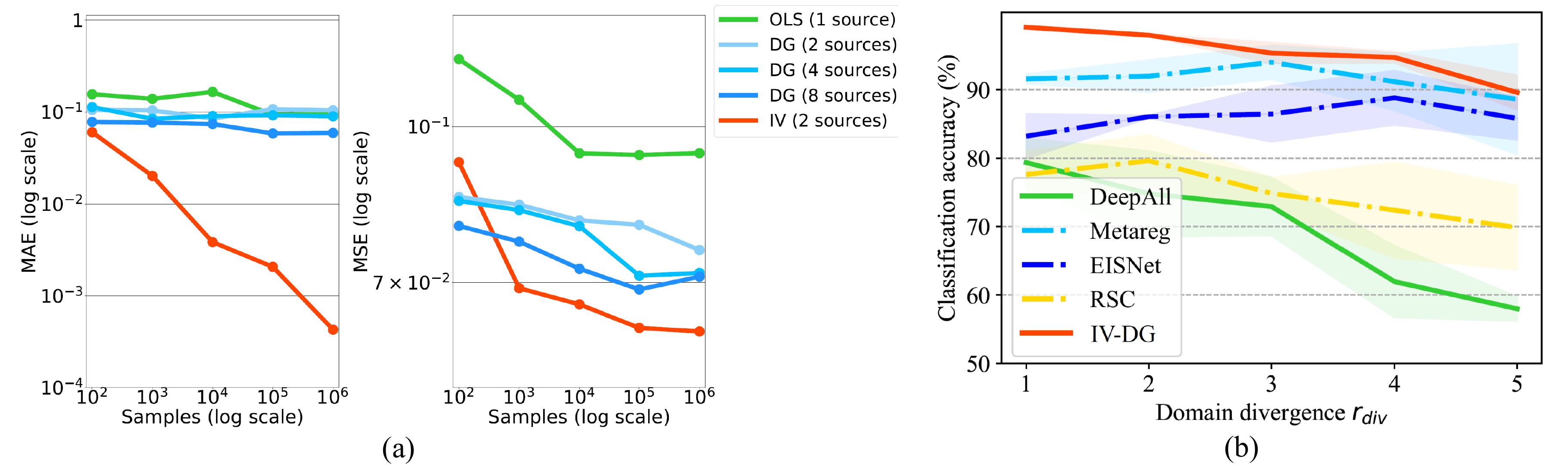}
    \caption{Simulation results in linear setting (a) for evaluating invariant relationship learning and target regression, and non-linear setting (b) for label prediction.}
    \label{linear-nonlinear}
\end{figure*}

\textbf{Linear simulations.}
We first evaluate the performance of invariant relationship estimation and target label prediction of IV method in linear domain generalization setting.
We sample variables for each domain $m$ with $F^{ivt}, F^{m} \sim \mathcal{N}(\mu_{f},1)$ and $e_{x}^{m}, e_{y}^{m} \sim \mathcal{N}(\mu_{e},0.1)$. We sample $\mu_{f}$ once from uniform distribution $\mathrm{Unif}(-1,1)$ and sample $\mu_{e}$ once from $\mathrm{Unif}(-0.1,0.1)$ for each domain, making the divergence in each domain be random. 
We first consider linear setting with one-dimensional variables. 
The DGP of Figure \ref{fig-dgp-all} is assumed as
\begin{equation}
\begin{aligned}
        X^{m}=&\phi_{m}\cdot F^{ivt}+\alpha_{m}\cdot F^{m}+e_{x}^{m}, \\
    Y^{m}=&\lambda_{ivt}\cdot X^{m}+\beta_{m}\cdot F^{m}+e_{y}^{m},
\end{aligned}
\end{equation}
where $\lambda_{ivt}$ is the invariant relationship that we are interested in. 
We sample $\phi_{m}$, $\lambda_{ivt}$ once from $\mathrm{Unif}(-1,1)$ and sample $\alpha_{m}$, $\beta_{m}$ once from $\mathrm{Unif}(-0.5,0.5)$ for each domain. Note that we let domain-invariant factor and relationship, i.e., $F^{ivt}$ and $\lambda_{ivt}$, be the same in all domains. In each run, we randomly generate 8 source domains for training and a target domain for test with 20,000 points in each domain. 
We run each method with linear regression, and report the MAE of domain-invariant relationship estimation, i.e., $\mathbb{E}[\hat{\lambda}_{ivt}-\lambda_{ivt}]$, and the MSE of the target domain label $Y^{t}$ prediction, i.e., $\mathbb{E}[(\hat{Y^{t}}-Y^{t})]^{2}$. 
We implement \textbf{OLS} method by training the model on one source domain. The general DG method is implemented by estimating the coefficient in each domain and average them to get a robust coefficient. \textbf{DG (n) sources} is denoted as the coefficient estimated in this way with $n$ sources. \textbf{IV} method only needs two sources, i.e., the input features of one is used as IV to estimate the relationship on another source domain. 
We plot the results in Figure \ref{linear-nonlinear} (a). Obviously, with the increase of sample size, IV method outperforms others in invariant relationship $\lambda_{ivt}$ estimation and target label prediction when only using two source datasets. Although more source datasets allow the general DG methods to eliminate the domain-specific bias, they are still fooled by the introduced bias in data. 

\textbf{Non-linear simulations.}
We further evaluate the performance of our IV-based method IV-DG in non-linear domain generalization setting.
Similar to the DGP in linear simulations, we sample variables $F^{ivt},F^{m}\sim\mathcal{N}(\mu_{f},1)$ and $e_{x}^{m},e_{y}^{m}\sim\mathcal{N}(\mu_{e},0.1)$. To evaluate the performance change with domain divergence, we sample $\mu_{f}$ from $\mathrm{Unif}(-r_{div},r_{div})$ for each domain, where a larger value of $r_{div}$ indicates the larger the domain divergence probably be. The DGP in non-linear setting is assumed as
\begin{equation}
    \begin{aligned}
        X^{m}=&\phi_{m}\cdot F^{ivt} + \alpha_{m}\cdot F^{m}+e_{x}^{m}, \\
        Y^{m}=&f_{ivt}(X^{m})+\beta_{m}\cdot F^{m}+e_{y}^{m}
    \end{aligned}
\end{equation}
The invariant relationship $\lambda_{ivt}$ is replaced with a non-linear function $f_{ivt}$, which is set to the absolute value function in the experiments. We set the dimensions of factor, i.e., $F^{ivt}$ and $F^{m}$, and input features, i.e., $X^{m}$, to 1500 and 600, respectively. 
We sample 10,000 data points each domain and divide them evenly into two classes with a threshold for the label $Y^{m}$. The goal is to accurately classify the target data by learning from 8 source datasets. We compare IV-DG with state-of-the-art DG methods (introduced in Section \ref{sec-rel}), i.e., Metareg \cite{balaji2018metareg}, EISNet \cite{Wang2020LearningFE}, and RSC \cite{HuangWXH20}.

All the methods are implemented by their public code, but their networks are replaced with 4 fully-connected layers with 600, 256, 128, and 64 units, respectively, for fair comparison. We use SGD optimizer with learning rate 0.01, and run 4000 iterations with batchsize 64. 
The results in Figure \ref{linear-nonlinear} (b) illustrates that IV-DG with IV-based two-stage method outperforms other state-of-the-art DG methods. It is worth mentioning that IV-DG only utilizes two source domains to train, while other methods have 8 sources. We attribute the significant performance of IV-DG to its domain-invariant relationship learning ability, which makes full use of the two sources to obtain the invariant part contained in the conditional distribution of the labels given the input features. Besides, we find that data augmentation based method, i.e., Metareg and EISNet, show the robustness to the domain divergence. It is may because these methods generate various data distributions, and the models trained on the novel data could be more robustness.

\subsection{Experiments on Real-World Datasets}
\begin{figure*}[t]
    \centering
    \includegraphics[trim={0cm 0cm 0cm 0cm},clip,width=0.24\columnwidth]{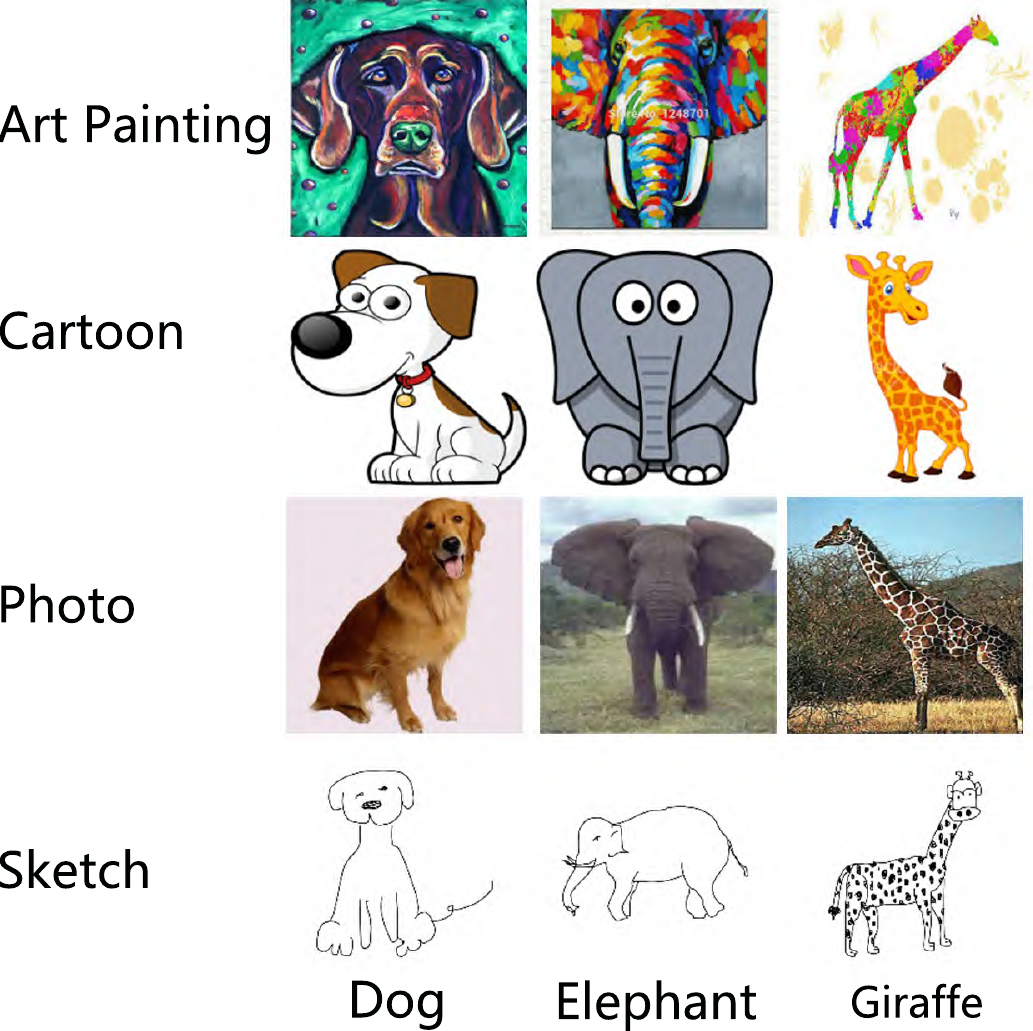}
    \includegraphics[trim={0cm 0cm 0cm 0cm},clip,width=0.24\columnwidth]{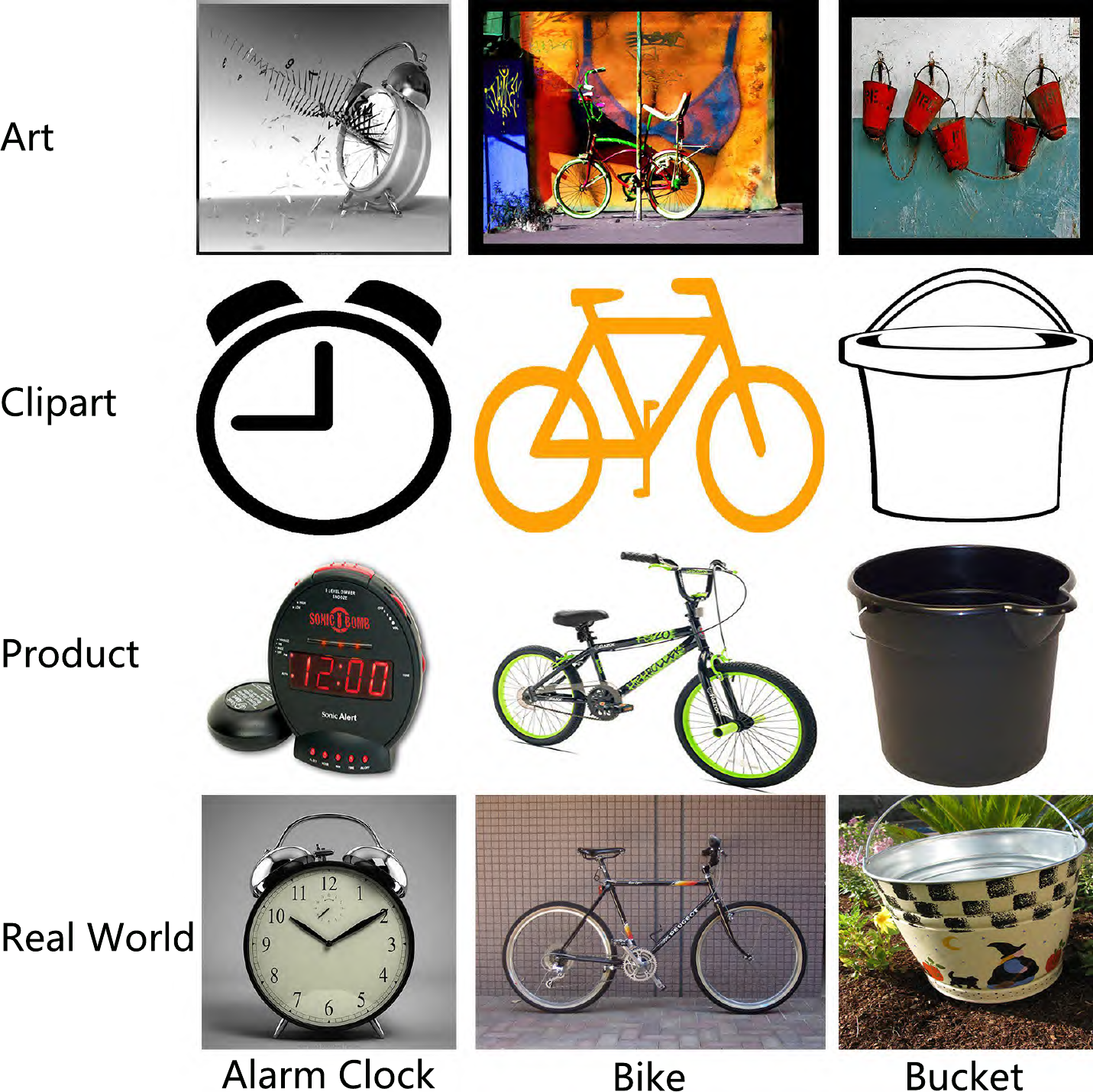}
    \includegraphics[trim={0cm 0cm 0cm 0cm},clip,width=0.24\columnwidth]{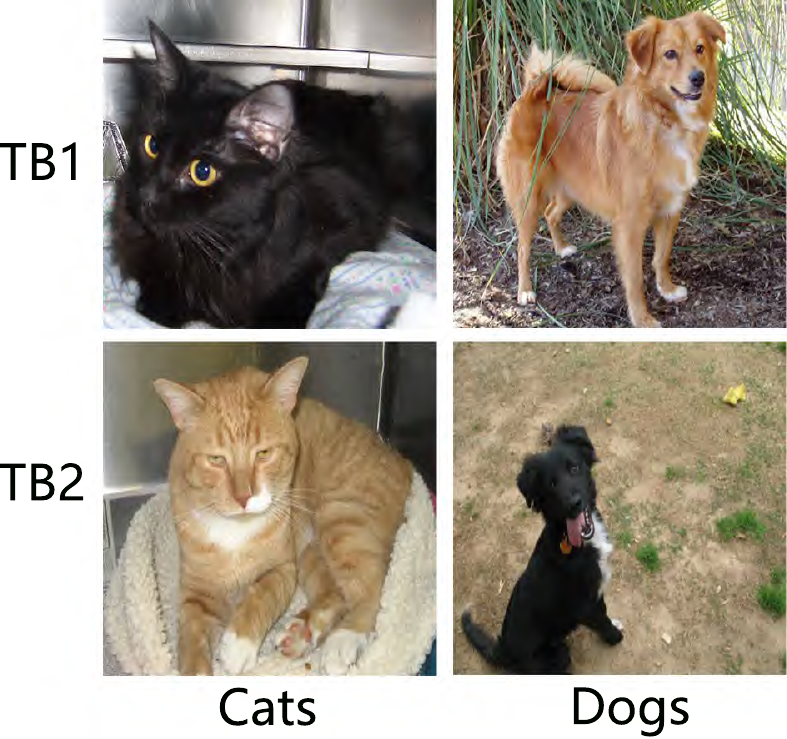}
    \includegraphics[trim={0cm 0cm 0cm 0cm},clip,width=0.24\columnwidth]{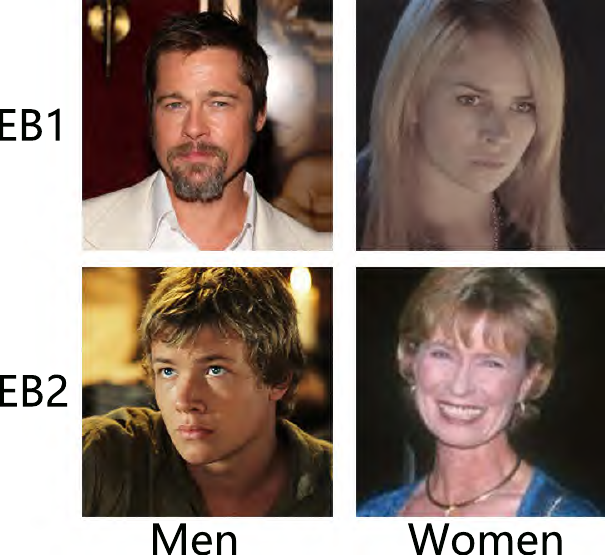}
    \caption{Example images of the adopted public datasets from left to right: PACS, Office-Home, Dogs and Cats, IMDB face. The former two datasets are used for the domain generalization task; and the latter two datasets are used for the unsupervised domain adaptation task.}
    \label{fig-dataset}
\end{figure*}

\textbf{Datasets and Implementations.}
We first conduct experiments on \textbf{PACS} \cite{li2017deeper}, which has 7 categories over 4 domains, that is, Art, Cartoon, Sketch, and Photo. 
Then we have \textbf{Office-Home} dataset \cite{venkateswara2017deep} that consists of 15,500 images of 65 categories over 4 domains, i.e., Art, Clipart, Product, and Real-World. 
Example images are shown in Figure \ref{fig-dataset}.
We follow the training and test split in previous works \cite{venkateswara2017deep, li2017deeper, Zhao2020DomainGV}, and perform leave-one-domain-out experiments, i.e., one domain is held out as the target domain for test. 
We follow \cite{Carlucci2019DomainGB, dou2019domain, HuangWXH20} by using the pretrained ResNet-18 \cite{he2016deep} network. We use SGD optimizer with learning rate 0.01 and batchsize 64. The epochs for the pretraining ($E^{pre}$) and the IV method ($E^{IV}$) are both set to 20. As one domain is chosen as the target domain, any of the rest domains can be used as $\mathcal{D}^{1}$, we use a held-out validation set, which is constructed from test domain by following the previous DG works \cite{Zhao2020DomainGV, zhou2021domain, zhou2020learning, HuangWXH20}, to choose the optimal $\mathcal{D}^{1}$ as well as the corresponding hyper-parameters of Eq. (\ref{equ-iv2}). 
We conduct the experiments with CPU Intel i7-8700K $\times$ 1 and GPU Nvidia RTX 3090 $\times$ 1. We run each experiment 3 times with random seed, and cite the results of other methods in their papers (note that some baseline methods are not in Table \ref{table-pacs} or Table \ref{table-office-home}
because their results are not reported in the corresponding paper).

Since when a domain is used as the target domain, any source could be treated as the $\mathcal{D}^{1}$, and other sources are used to learn the conditional distribution of $X^{1}$. Therefore, we first set all the weights (hyper-parameters) $\alpha$ to 1, and conduct different source combination experiments for PACS (Table \ref{table-comb-pacs}) and Office-Home (Table \ref{table-comb-office-home}) datasets.
From Table \ref{table-comb-pacs} and Table \ref{table-comb-office-home}, we observe that different choices for the first domain would not have a significant impact on the results, which shows the robustness of our method.
After we have the best domain combinations, we then conduct weight combination experiments on PACS (Table \ref{table-weight-pacs}) and Office-Home (Table \ref{table-weight-office-home}) datasets.
Finally, 
we use the ``target-$\mathcal{D}^{1}$'' combinations ``Art-Photo'', ``Cartoon-Photo'', ``Photo-Art'', ``Sketch-Photo'' with weights $\alpha_{1}=1.25, \alpha_{2}=0.75$ for PACS dataset; and use ``Art-Clipart'', ``Clipart-Art'', ``Product-Art'', ``Real-World-Art'' with weights $\alpha_{1}=1.5, \alpha_{2}=0.5$ for Office-Home datasets. 

\begin{table}[t]
\caption{Results (\%) for domain generalization on PACS dataset.}
    \label{table-pacs}
    \centering
\resizebox{1\linewidth}{!}{
\renewcommand\tabcolsep{5.0pt}
\begin{tabular}{lcccc|c}
\toprule
Methods & Art & Cartoon & Photo & Sketch & Average \\
\midrule
DeepAll \cite{Carlucci2019DomainGB} & 78.96 & 72.93 & 96.28 & 70.59 & 79.94 \\
JiGen \cite{Carlucci2019DomainGB} & 79.42 & 75.25 & 96.03 & 71.35 & 80.51 \\
MASF \cite{dou2019domain} & 80.29 & 77.17 & 94.99 & 71.69 & 81.04 \\
DGER \cite{Zhao2020DomainGV} & 80.70 & 76.40 & 96.65 & 71.77 & 81.38 \\
Epi-FCR \cite{Li2019EpisodicTF} & 82.1 & 77.0 & 93.9 & 73.0 & 81.5 \\
MMLD \cite{Matsuura2020DomainGU} & 81.28 & 77.16 & 96.09 & 72.29 & 81.83 \\
EISNet \cite{Wang2020LearningFE} & 81.89 & 76.44 & 95.93 & 74.33 & 82.15 \\
L2A-OT \cite{zhou2020learning} & 83.3 & 78.2 & 96.2 & 73.6 & 82.8 \\
DDAIG \cite{zhou2020deep} & \textbf{84.2} & 78.1 & 95.3 & 74.7 & 83.1 \\
IRM \cite{arjovsky2019invariant} & 82.5 & 79.0 & 96.7 & 74.4 & 82.9 \\
StableNet \cite{zhang2021deep} & 80.16 & 74.15 & 94.24 & 70.10 & 79.66 \\

\midrule
IV-DG w/o IV & 79.40 $\pm$ 0.10 & 76.93 $\pm$ 0.09 & 95.75 $\pm$ 0.10 & 74.44 $\pm$ 0.07 & 81.63 $\pm$ 0.03 \\
IV-DG w/o pre & 81.95 $\pm$ 0.25 & 77.55 $\pm$ 0.31 & 96.64 $\pm$ 0.34 & 75.65 $\pm$ 0.10 & 82.95 $\pm$ 0.14 \\
IV-DG & 83.36 $\pm$ 0.70 & \textbf{78.76} $\pm$ \textbf{0.08} & \textbf{96.87} $\pm$ \textbf{0.18} & \textbf{78.68} $\pm$ \textbf{0.96} & \textbf{84.42} $\pm$ \textbf{0.11} 
\\
\bottomrule
\end{tabular}}
\end{table}

\begin{table}[t]
\caption{Results (\%) for domain generalization on Office-Home dataset.}
    \label{table-office-home}
    \centering
\resizebox{1\linewidth}{!}{
\begin{tabular}{lcccc|c}
\toprule
Methods & Art & Clipart & Product & Real-World & Average \\
\midrule
DeepAll \cite{Carlucci2019DomainGB} & 52.15 & 45.86 & 70.86 & 73.15 & 60.51 \\
JiGen \cite{Carlucci2019DomainGB} & 53.04 & 47.51 & 71.47 & 72.79 & 61.20 \\
DSON \cite{Seo2020LearningTO} & 59.37 & 44.70 & 71.84 & 74.68 & 62.90 \\ 	
RSC \cite{HuangWXH20} & 58.42 & \textbf{47.90} & 71.63 & 74.54 & 63.12 \\
\midrule
IV-DG w/o IV & 55.53 $\pm$ 0.21 & 45.92 $\pm$ 0.50 & 71.64 $\pm$ 0.35 & 74.49 $\pm$ 0.05 & 61.90 $\pm$ 0.20 \\
IV-DG w/o pre & 59.30 $\pm$ 0.06 & 47.65 $\pm$ 0.30 & 72.03 $\pm$ 0.57 & 75.55 $\pm$ 0.24 & 63.63 $\pm$ 0.11 \\
IV-DG & \textbf{60.40} $\pm$ \textbf{0.26} & 47.73 $\pm$ 0.28 & \textbf{72.63} $\pm$ \textbf{0.18} & \textbf{76.14} $\pm$ \textbf{0.10} & \textbf{64.23} $\pm$ \textbf{0.09} \\
\bottomrule
\end{tabular}}
\end{table}

\begin{table}[t]
\caption{Results (\%) of different combinations for domain generalization on PACS dataset.}
    \label{table-comb-pacs}
    \centering
\resizebox{0.7\linewidth}{!}{
\begin{tabular}{c|cccc}
\toprule
$\mathcal{D}^{1}$ $\backslash$ Target & Art & Cartoon & Photo & Sketch \\
\midrule
Art & - & 78.10 $\pm$0.37 & \textbf{97.17 $\pm$ 0.12} & 76.91 $\pm$ 0.03 \\
Cartoon & 82.21$\pm$0.97 & - & 96.75 $\pm$ 0.17 & 76.78 $\pm$ 0.98 \\
Photo & \textbf{83.77 $\pm$ 0.57} & \textbf{78.34 $\pm$ 0.58} & - & \textbf{77.48 $\pm$ 0.32} \\
Sketch & 81.46 $\pm$ 0.10 & 78.20 $\pm$0.76 & 97.01 $\pm$ 0.27  & - \\
\bottomrule
\end{tabular}}
\end{table}

\begin{table}[t]
\caption{Results (\%) of different combinations for domain generalization on Office-Home dataset.}
    \label{table-comb-office-home}
    \centering
\resizebox{0.7\linewidth}{!}{
\begin{tabular}{c|cccc}
\toprule
$\mathcal{D}^{1}$ $\backslash$ Target & Art & Clipart & Product & Real-World \\
\midrule
Art & - & \textbf{45.71 $\pm$ 0.20} & \textbf{72.31 $\pm$ 0.22} & \textbf{76.88 $\pm$ 0.08} \\
Clipart & \textbf{60.89 $\pm$ 0.17} & - & 72.21 $\pm$ 0.05 & 76.88 $\pm$ 0.12 \\
Product & 60.41 $\pm$ 0.20 & 45.10 $\pm$ 0.53 & - & 76.83 $\pm$ 0.14 \\
Real-World & 60.64 $\pm$ 0.29 & 45.65 $\pm$ 0.23 & 72.30 $\pm$ 0.41  & - \\
\bottomrule
\end{tabular}}
\end{table}

\begin{table}[t]
\caption{Results (\%) with different weights for domain generalization on PACS dataset.}
    \label{table-weight-pacs}
    \centering
\resizebox{0.9\linewidth}{!}{
\begin{tabular}{cc|cccc|c}
\toprule
$\alpha_{1}$ & $\alpha_{2}$ & Art & Cartoon & Photo & Sketch & Average \\
\midrule
0 & 2 & 82.68 $\pm$ 0.23 & 78.25 $\pm$ 0.15 & 97.15 $\pm$ 0.18 & 77.43 $\pm$ 0.29 & 83.88 $\pm$ 0.88 \\
0.25 & 1.75 & 82.33 $\pm$ 0.30 & 79.15 $\pm$ 0.57 & 97.07 $\pm$ 0.12 & 78.16 $\pm$ 0.64 & 84.18 $\pm$ 0.26 \\
0.5 & 1.5 & 82.21 $\pm$ 0.75 & 78.19 $\pm$ 0.50 & 97.11 $\pm$ 0.07 & 77.51 $\pm$ 0.96 & 83.75 $\pm$ 0.09 \\
0.75 & 1.25 & 82.60 $\pm$ 0.88 & 78.88 $\pm$ 0.89 & 97.09 $\pm$ 0.03 & 78.65 $\pm$ 0.71 & 84.31 $\pm$ 0.62 \\
1 & 1 & \textbf{83.77} $\pm$ \textbf{0.57} & 78.34 $\pm$ 0.58 & 97.17 $\pm$ 0.12 & 77.48 $\pm$ 0.32 & 84.19 $\pm$ 0.25 \\
1.25 & 0.75 & 83.36 $\pm$ 0.70 & 78.76 $\pm$ 0.08 & 96.87 $\pm$ 0.18 & \textbf{78.68} $\pm$ \textbf{0.96} & \textbf{84.42} $\pm$ \textbf{0.11} \\
1.5 & 0.5 & 81.89 $\pm$ 0.08 & 78.60 $\pm$ 0.29 & \textbf{97.35} $\pm$ \textbf{0.12} & 78.20 $\pm$ 1.02 & 84.01 $\pm$ 0.18  \\
1.75 & 0.25 & 81.98 $\pm$ 0.15 & \textbf{79.17} $\pm$ \textbf{0.73} & 96.87 $\pm$ 0.38 & 78.53 $\pm$ 0.14 & 84.14 $\pm$ 0.10 \\
2 & 0 & 82.14 $\pm$ 0.17 & 78.22 $\pm$ 0.10 & 97.05 $\pm$ 0.12 & 77.46 $\pm$ 1.58 & 83.72 $\pm$ 0.38 \\
\bottomrule
\end{tabular}}
\end{table}

\begin{table}[t]
\caption{Results (\%) with different weights for domain generalization on Office-Home dataset.}
    \label{table-weight-office-home}
    \centering
\resizebox{0.9\linewidth}{!}{
\begin{tabular}{cc|cccc|c}
\toprule
$\alpha_{1}$ & $\alpha_{2}$ & Art & Clipart & Product & Real-World & Average \\
\midrule
0 & 2 & 60.63 $\pm$ 0.25 & 47.40 $\pm$ 0.08 & 72.51 $\pm$ 0.08 & 76.12 $\pm$ 0.59 & 64.16 $\pm$ 0.10 \\
0.25 & 1.75 & 60.71 $\pm$ 0.13 & 46.48 $\pm$ 0.26 & 72.59 $\pm$ 0.08 & 76.93 $\pm$ 0.17 & 64.18 $\pm$ 0.02 \\
0.5 & 1.5 & 60.79 $\pm$ 0.11 & 46.18 $\pm$ 0.16 & 72.62 $\pm$ 0.13 & 76.10 $\pm$ 0.17 & 63.92 $\pm$ 0.04 \\
0.75 & 1.25 & 60.53 $\pm$ 0.09 & 46.36 $\pm$ 0.36 & 72.60 $\pm$ 0.14 & 76.69 $\pm$ 0.11 & 64.05 $\pm$ 0.08 \\
1 & 1 & 60.89 $\pm$ 0.17 & 45.71 $\pm$ 0.20 & 72.31 $\pm$ 0.22 & 76.88 $\pm$ 0.08 & 63.95 $\pm$ 0.05 \\
1.25 & 0.75 & 60.90 $\pm$ 0.39 & 46.20 $\pm$ 0.35 & 72.54 $\pm$ 0.23 & \textbf{77.06} $\pm$ \textbf{0.25} & 64.17 $\pm$ 0.10 \\
1.5 & 0.5 & 60.40 $\pm$ 0.26 & \textbf{47.73} $\pm$ \textbf{0.28} & \textbf{72.63} $\pm$ \textbf{0.18} & 76.14 $\pm$ 0.10 & \textbf{64.23} $\pm$ \textbf{0.09} \\
1.75 & 0.25 & 60.58 $\pm$ 0.20 & 46.48 $\pm$ 0.21 & 72.54 $\pm$ 0.14 & 76.89 $\pm$ 0.34 &  64.12 $\pm$ 0.11 \\
2 & 0 & \textbf{60.95} $\pm$ \textbf{0.15} & 46.28 $\pm$ 0.11 & 72.44 $\pm$ 0.04 & 76.88 $\pm$ 0.21 & 64.14 $\pm$ 0.04 \\
\bottomrule
\end{tabular}}
\end{table}

\begin{figure*}[t]
    \centering
    \includegraphics[trim={0cm 0cm 0cm 0cm},clip,width=0.99\columnwidth]{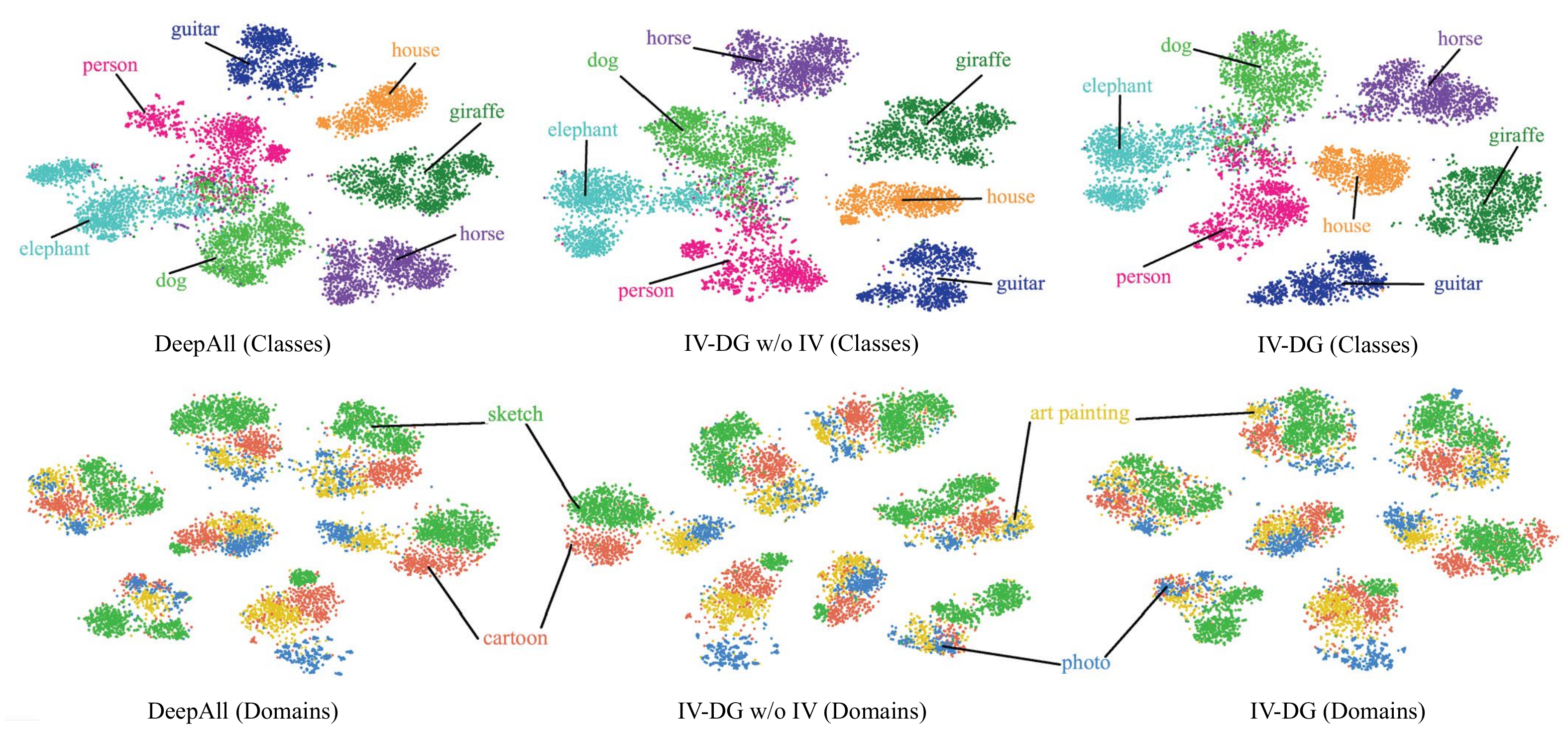}
    \caption{T-SNE visulazation of the learned feature representations of DeepAll, IV-DG w/o IV, and IV-DG, on PACS dataset. Different colors in the above and below sub-figures represent different classes and domains, respectively. The points gather separately for classes and compactly for domains indicates the learned feature representations are more discriminative and domain-agnostic, respectively.}
    \label{fig-tsne}
\end{figure*}

\begin{figure*}[t]
    \centering
    \includegraphics[trim={0cm 0cm 0cm 0cm},clip,width=0.99\columnwidth]{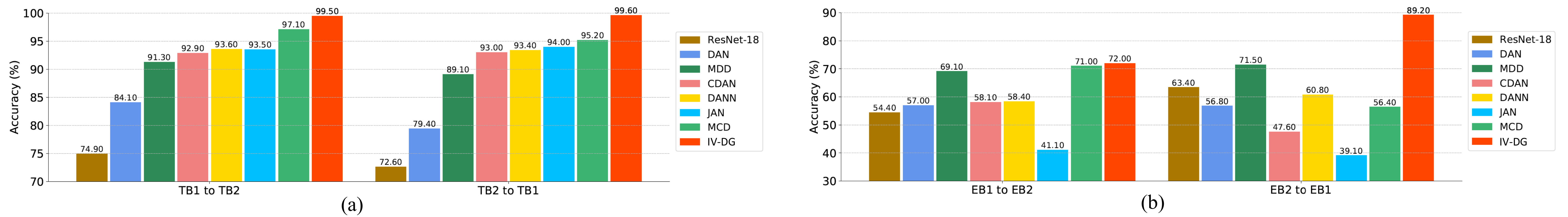}
    \caption{Results for unsupervised domain adaptation task on Dogs and Cats (a) and IMDB face (b) datasets.}
    \label{fig-uda}
\end{figure*}



\textbf{Results.} 
Table \ref{table-pacs} and Table \ref{table-office-home} report the results on PACS and Office-Home datasets, respectively. 
Note that the \textbf{DeepAll} method is implemented by training the model with general supervised learning on the aggregation of all the source datasets.
We first find that IV-DG outperforms other methods on both datasets by performing the best on most of the DG sub-tasks and achieving the highest averaged accuracy. We attribute it to that IV-DG learns to capture the invariant part (relationship) contained in the conditional distribution for better model generalization. 
We let IV-DG discard the IV method and pretraining as \textbf{w/o IV} and \textbf{w/o pre}, respectively in Table \ref{table-pacs} and Table \ref{table-office-home}. It shows that each part is important for IV-DG to yield significant performance, especially the IV method. It is may because the pretrainig initialize the discriminability of the feature extractor for better conditional distribution estimation, and the IV method helps model learn domain-invariant relationship by debiasing the classifier. We plot the t-SNE feature visualization in Figure \ref{fig-tsne}. It indicates that IV method helps IV-DG to learn discriminative and domain-invaraint feature during the IV-based two-stage process by separating the features of different classes while aggregating the features of different domains.

\subsection{Experiments on Biased Data}\label{sec:biased-data}
We also evaluate IV-DG on the unsupervised domain adaptation (UDA) task where IV-DG uses the input features of the target dataset as IVs to learn the invariant relationship with the given source dataset. 
We adopt two biased datasets for this task. The first is \textbf{Dogs and Cats} \cite{kim2019learning}, where TB1 domain contains bright dogs and dark cats; but TB2 domain contains dark dogs and bright cats. The second is \textbf{IMDB face} dataset \cite{kim2019learning}. Women in a domain EB1 aged 0-29 and in another domain EB2 aged 40+; but men in EB1 aged 40+ and in EB2 aged 0-29. There is clear bias between the domains in the two datasets, which challenges the methods to learn stable relationship between the images and labels. We compare IV-DG with representative DA approaches, DAN \cite{long2015learning}, DANN \cite{ganin2015unsupervised}, JAN \cite{long2017deep}, MDD \cite{Zhang2019BridgingTA}, CDAN \cite{long2018conditional}, MCD \cite{saito2018maximum}. All the experiments are implemented using the same training setting for fair comparison.
Following \cite{long2018conditional,xu2019larger,Liang2020DoWR}, We employ the pre-trained ResNet-50 \cite{he2016deep} as the feature extractor, where the last layer is replaced by one FC layer with 256 units. Classifier is a FC layer put after feature extractor for classification. We train each method through back-propagation by SGD with batch-size 64, learning rate 0.01, momentum 0.9, and weight decay 0.001. Each method are run 10 epochs on Dogs and Cats dataset and 5 epochs on IMDB face dataset for fair comparison. 
Results in Figure \ref{fig-uda} show that IV-DG performs much better than others on the two challenging biased datasets. Moreover, we find that IV-DG achieves significant improvement on IMDB face dataset. It is probably because IV method needs sufficient samples to obtain invariant relationship (see Figure \ref{linear-nonlinear}), and IMDB face is a large dataset with 460,723 images.

\section{Conclusions}\label{sec-con}
In this paper, we first give a causal view on the domain generalization problem, and then propose to learn domain-invariant relationship with instrumental variable via an IV-based two-stage method. Extensive experiments show the significant performance of our method. Our paper benefits the research of domain generalization and may not have negative impact of society to our knowledge. 

Despite the great performance achieved by our method, there are some limitations. 
First, our method is based on the assumed causal graph. Some assumptions, e.g., $F^{m}$ is the domain-specific factor changed with the background but uncorrelated to the invariant factor of the recognized animals, may not hold in real-world scenarios. For example, lambs with their size and limbs probably in grassland.
Second, since our theoretical analyses relies on the assumption of linearity and additivity, hence it may be not leading to stable prediction on out-of-distribution target data when the data generating process of domain-invariant factor $F^{ivt}$, domain-specific factor $F^m$, and error term $e_x^m$, $e_y^m$ is a highly non-linear complex function \cite{rothenhausler2021anchor, lin2022orphicx}. For example, the bias of domain-specific factor (confounder) $F^m$, which causes the distribution shift, may not be completely removed if $F^m$ is connected with $f(X^m)$ via an unknown non-linear function. In the future work, we aim to use more moderate assumptions to build the model, which could achieve better performance.

\begin{acks}
This work was supported in part by the National Key Research and Development Project (No. 2022YFC2504605), National Natural Science Foundation of China (62006207, U20A20387, 62037001), Young Elite Scientists Sponsorship Program by CAST (2021QNRC001), Zhejiang Provincial Natural Science Foundation of China (No. LZ22F020012), Major Technological Innovation Project of Hangzhou (No. 2022AIZD0147), Zhejiang Province Natural Science Foundation (LQ21F020020), Project by Shanghai AI Laboratory (P22KS00111), Program of Zhejiang Province Science and Technology (2022C01044), the StarryNight Science Fund of Zhejiang University Shanghai Institute for Advanced Study (SN-ZJU-SIAS-0010), and the Fundamental Research Funds for the Central Universities (226-2022-00142, 226-2022-00051).
\end{acks}

\bibliographystyle{ACM-Reference-Format}
\bibliography{sample-base}

\appendix

\end{document}